\documentclass[11pt,a4paper]{article}
\usepackage[T1]{fontenc}
\usepackage[utf8]{inputenc}
\usepackage{mathptmx}
\usepackage[scaled=0.92]{helvet}
\usepackage{courier}
\usepackage{textcomp}
\usepackage{amssymb}
\usepackage[margin=2.4cm]{geometry}
\usepackage{microtype}
\usepackage{graphicx}
\usepackage{longtable,booktabs,array,calc}
\usepackage{xcolor}
\usepackage[font=small,labelformat=empty,skip=6pt]{caption}
\usepackage{newunicodechar}
\newunicodechar{Δ}{\ensuremath{\Delta}}
\newunicodechar{→}{\ensuremath{\rightarrow}}
\newunicodechar{≥}{\ensuremath{\geq}}
\usepackage[hidelinks]{hyperref}
\makeatletter
\def\fps@figure{tbp}
\makeatother
\begin{document}

\begin{center}
{\LARGE\bfseries MuninAI: A Self-Hosted Agentic Framework\\[2pt] for Independent Research Groups\par}
\vspace{1.2em}
{\large Markus Fischer\textsuperscript{1,2,*}, Jens Meiler\textsuperscript{1,4,5,6}, Matthias Elgeti\textsuperscript{1,2,3,*}\par}
\vspace{0.8em}
{\small
\textsuperscript{1}Institute for Drug Discovery, Leipzig University Medical School, Leipzig, Germany\\
\textsuperscript{2}Institute for Medical Physics and Biophysics, Leipzig University Medical School, Leipzig, Germany\\
\textsuperscript{3}Integrative Center for Bioinformatics, Leipzig University, Leipzig, Germany\\
\textsuperscript{4}Center for Scalable Data Analytics and Artificial Intelligence (ScaDS.AI), Dresden/Leipzig, Germany\\
\textsuperscript{5}School of Embedded Composite Artificial Intelligence (SECAI), Cooperation of University Leipzig and TU Dresden, Dresden/Leipzig, Germany\\
\textsuperscript{6}Faculty of Mathematics and Informatics, Faculty of Chemistry, Leipzig University, Leipzig, Germany\\[6pt]
\textsuperscript{*}Corresponding authors: \href{mailto:markus.fischer@medizin.uni-leipzig.de}{markus.fischer@medizin.uni-leipzig.de}, \href{mailto:matthias.elgeti@medizin.uni-leipzig.de}{matthias.elgeti@medizin.uni-leipzig.de}\\[4pt]
ORCID: Markus Fischer 0000-0002-7057-2309; Jens Meiler 0000-0001-8945-193X; Matthias Elgeti 0000-0002-0550-4852\\[4pt]
Code, evaluation suite and all committed scorecards: \url{https://github.com/elgetilab/munin}\\
(archived release v1.0.0: \url{https://doi.org/10.5281/zenodo.23191404}). Preprint version 1, 16 October 2026.\par}
\end{center}
\vspace{0.5em}

\begin{abstract}
Research groups that want a language model grounded in their own literature face a choice between hosted services that receive their unpublished work and self-hosted deployments that keep the data local but give the model no access to the corpus. Munin is an open-source framework for a self-hosted scientific assistant that indexes a group\textquotesingle s corpus and its citation graph, and wraps whichever open-weight model the group can serve in an agent harness with specialised reading, search, and computation tools. We claim that the harness, rather than the specific model, is what makes the system useful on domain literature, and we ship the evaluation suite that tests this claim with the software.

On 199 paired literature questions, the harness more than doubles the accuracy of the base model, from 0.387 to 0.874 (+0.487, 95\% CI {[}0.407, 0.568{]}), while naive retrieve-then-answer scores below the unaided model on every backbone tested. Under the harness, roughly twice as many answer claims trace to a passage the model read itself as with fixed top-5 retrieval. When the source paper for a question is removed from the corpus, the system abstains on about nine in ten of the questions it had answered correctly on the Qwen backbones and eight in ten on gpt-oss-20b, and it produces no fabricated local citation across 300 fabricated references.

The pattern holds on three backbones from two developers, including a 20-billion-parameter model that runs on a single consumer-grade GPU. With all outbound network access disabled, both Qwen backbones still exceed the published PaperQA2 accuracy. The evaluation suite runs as a single command, so an adopting group can re-run it on its own corpus and model.
\end{abstract}

\section*{1 Introduction}\label{introduction}

Every research group accumulates a body of literature over years: papers they cite repeatedly, foundational references, preprints shared by collaborators, and results that have not yet been published. Locating a specific value, figure, or methodological detail within that collection typically depends on individual scientists remembering which document contains it. Citation managers organise bibliographies during the writing process, but extracting quantitative data from figures and tables remains a manual task.

Commercial language-model services and agentic search tools have made it straightforward to query the open web, however a group\textquotesingle s internal corpus sits outside that reach: it includes licensed articles whose redistribution is restricted, unpublished manuscript drafts, instrument-specific datasets, and internal reports that no public index covers. What a public model can answer from the open literature and what the group actually needs to retrieve from its own files are therefore different questions, and the gap between them widens as the corpus grows.

We present Munin, a framework for running an agentic research assistant on a group\textquotesingle s own hardware. A group ingests its own papers and obtains an assistant that answers from that private collection. The framework is model-agnostic. Any serving stack that exposes an OpenAI-compatible API is a valid inference target, and the system scales down to a single GPU. Beyond the assistant itself, Munin ships a frozen evaluation suite so that an adopting group can verify the performance claims on its own corpus rather than relying on our reported numbers.

The system has two halves, deliberately decoupled so that a group may deploy either independently. The backend holds the agent loop, a knowledge base (vector index, citation graph, PDF parser, ingestion pipeline), and a client for the inference endpoint. It runs entirely on the institution\textquotesingle s infrastructure. The frontend handles authentication, file upload, and the chat interface. As an added security layer, the frontend reaches the backend through an outbound tunnel so that the machine storing the corpus accepts no inbound connection. A group that already runs its own interface can call the backend API directly and deploy only that half. The language model itself is a configuration parameter and not a hard dependency.

The corpus is whatever the group uploads. On ingestion, documents are labelled by contributing group into sub-corpora, and clustered into topics, so that users can scope queries to a relevant subset. Because the correctness of the agent loop depends jointly on the corpus content, the chosen model, and the domain of the questions, we ship the benchmarking suite alongside the framework. An adopting group can re-run the full evaluation after any change to model, encoder, corpus, or harness, and compare against the reference numbers we report for three model configurations (gpt-oss-20b, Qwen3.6, Qwen3.8). We also include ablation baselines: the same questions answered by the bare model with no retrieval, and by naive top-k retrieval with no agentic loop. The difference between those baselines and the full system isolates the contribution of the iterative tool-use loop.

Our contributions are: (1) Munin, a self-hosted agentic framework consisting of a multi-step tool-use loop over 45 tools, an ingestion and corpus-quality pipeline, a vector-and-graph knowledge base, and a deployment topology in which the corpus host accepts no inbound traffic; (2) a frozen evaluation suite that an adopting group runs to establish its own baselines and to re-measure after any component change; (3) harness-value measurements across three model configurations, with per-question results released for inspection; (4) a corpus-grounding test that yields DOI-level ground truth by construction, requiring no human annotation and (5) egress level treated as a first-class experimental variable. We also documented our development record in the Appendix for other developers to use.

\section*{2 Related Work}\label{related-work}

\subsection*{2.1 Agentic systems for scientific literature}\label{agentic-systems-for-scientific-literature}

PaperQA2 {[}1{]} is the closest system precedent and the source of the LitQA2 benchmark used here. It reports accuracy 0.660 ± 0.012 and precision 0.852 ± 0.011 on LitQA2, against a human-expert accuracy of 0.677 ± 0.119 (n = 9) and precision 0.738 ± 0.096, using definitions identical to ours: accuracy is correct answers over all questions asked, precision is correct answers over questions answered. Its predecessor {[}2{]} and the Aviary environment framework {[}3{]} establish the retrieve-then-read-full-text pattern that our \emph{source} agent also implements.

There are two other open-source systems that are similar to Munin, that pair a large scientific corpus with a service layer around it. Ai2 Scholar QA {[}38{]} answers open-ended scientific questions over a Semantic-Scholar-indexed collection of roughly 100 million abstracts and 11.7 million full texts. A cross-encoder re-ranks the retrieved passages, a language model extracts and clusters quotes by sub-topic, and the output is a sectioned report with inline citations. OpenScholar {[}39{]} takes a different approach, training an 8B retrieval-augmented model over a 45-million-paper index for the same task and introducing ScholarQABench, on which it reports that a frontier model without retrieval fabricates the majority of its cited papers on biomedical questions while the grounded model does not. Both systems are served as hosted services over a global corpus. Neither indexes a group\textquotesingle s own unpublished collection, runs on the group\textquotesingle s hardware, or measures how the system behaves when the corpus does not contain the answer, which is the gap we want to address with Munin.

Most commonly, groups would self-host a chat frontend over a local inference server. This can be achieved with Open WebUI {[}40{]}, LibreChat {[}41{]}, and AnythingLLM {[}42{]}, which are the most widely deployed options. They keep data on premises, and AnythingLLM ships a general document-retrieval pipeline, but all three treat the model as a black box behind a chat window. None maintains a citation graph, reads full text before answering, routes a query to a not-in-corpus path, or includes an evaluation suite that a deploying group can re-run after a configuration change. Because these systems have no peer-reviewed description, we cite them as software rather than as published methods. The result in Section 5.1.1, that a fixed top-k context scores below the bare model on every backbone tested, is a direct measurement of the retrieval design these frontends implement.

AstaBench {[}22{]} is the closest precedent for rigorous agent benchmarking. They address over 2,400 problems across eleven sub-benchmarks in four categories (literature understanding, code and execution, data analysis, and end-to-end discovery), scored on a leaderboard that plots accuracy against cost under a standardised tool budget rather than ranking by accuracy alone. We adopt this cost-aware presentation, reporting wall-clock and tool calls alongside every accuracy figure. Three of their findings are directly relevant for our framework: First, their purpose-built Paper Finder agent roughly doubles the accuracy compared to a generic ReAct (a reason-act-observe loop) scaffold on paper-finding and gains 15\% on LitQA2-FullText-Search. Note that their baseline is a tool-using loop while our bare arm has no tools, making theirs the stricter comparison. The analogue in our framework is the agentic to RAG gap.

Second, and more importantly, it reports that a stronger base model does not uniformly help. The effect splits by agent type: gpt-5 gained roughly 3\% over o3 when driving a generic ReAct agent, but three of their specialised agents (Scholar QA, DataVoyager, and Code) each performed worse under gpt-5 than under the models they were built around. The authors speculate that gpt-5 was post-trained for ReAct-style control. A harness tuned around one model\textquotesingle s tool-calling behaviour can worsen the performance if used with another model, and a specialised harness is the exposed case. Munin is a specialised harness, not a ReAct loop. This is addressed in section 5.4, where we measure our harness performance against three models independently.

Third, their results calibrate what an open-weight system can currently be expected to achieve. The best open-weight configuration they evaluated reached 11.1\% on the cross-category overall score, against 53.0\% for their own Asta v0, which routes each problem to one of five task-specific agents spanning multiple language models. Because that aggregate includes code execution and data analysis, categories outside Munin\textquotesingle s scope, it isn't an immediate benchmark for us to compare against. However, it indicates that open-weight agents are not yet competitive across the full scientific workflow, whatever they achieve on a single task over a private corpus. Literature understanding was the most mature of their four categories, and the one in which our work sits.

PaperArena{[}23{]} evaluates tool-augmented agentic reasoning over scientific literature, requiring agents to synthesise information across multiple publications rather than extract it from a single source. Its best result, Gemini 2.5 Pro in a multi-agent configuration, reaches 38.78\% accuracy averaged over three answer formats and three difficulty tiers, falling to 18.47\% on the hard subset; PhD-level experts achieve 83.50\% on the same questions. Because the task requires cross-paper integration over a corpus of open-access AI conference papers, it is a harder task type than the single-source questions we report in Section 5.1.

AutoResearchBench {[}24{]} decomposes literature discovery into two tasks: Deep Research (locate a single target paper, scored by accuracy) and Wide Research (collect a relevant set, scored by IoU (Intersection over Union) against a gold set). Flagship models peak at 9.39\% and 9.31\% respectively, with many strong baselines below 5\%. Due to our much smaller models, we decided against evaluating against that benchmark, as it is unlikely that we would get any significant results.

LitSearch {[}4{]} contributes 597 natural-language queries over a 64,183-paper corpus with binary relevance labels and a mean of 1.07 relevant papers per query. It is the public benchmark whose task profile most closely matches our system\textquotesingle s intended use, and its near-single-relevance property is directly relevant to the citation re-ranking result in Section 5.5.

\subsection*{2.2 Abstention and selective prediction}\label{abstention-and-selective-prediction}

KnowOrNot {[}9{]} provides a library and methodology for evaluating out-of-knowledge-base robustness in retrieval-augmented systems. It measures whether a RAG system abstains when a question falls outside its knowledge base. It generates evaluation data without the need for annotated datasets. Its withholding mechanism is leave-one-out over a static policy corpus, and it uses a language model as judge only to classify whether a response constitutes an abstention. We ship a similar system focused on DOI-level ground truth described in section 4.6.

Sufficient Context {[}10{]} introduces a post-hoc classifier for whether already-retrieved context suffices to answer, and reports that models answer rather than abstain when it does not. We observe the same failure in our RAG baseline (Section 5.1.1) but address it architecturally rather than with a classifier: retrieval is iterative, so the loop can recognise insufficient evidence and retrieve more, and the reading agent is bound by a quote-or-abstain contract that requires a supporting sentence before an answer is emitted. In a static pipeline, all context is retrieved before the model begins to answer, and a classifier sits at that boundary. In an agentic loop, retrieval and answering are interleaved, so no such boundary exists at which to place the classifier.

RefusalBench {[}11{]} rewrites passages to be flawed, in six ways (ambiguous, contradictory, missing the key fact, falsely premised, wrongly scaled, or speculative), each at three severity levels, and checks whether the model notices. The document always remains and only its content is damaged, whereas we remove the paper from the corpus outright. They test whether a model is fooled by bad evidence; we test whether it notices when evidence is absent.\\
AbstentionBench {[}12{]} collects roughly 20 datasets and over 35,000 questions that have no answer at all, because they rest on a false premise, are too vague to answer, or ask something nobody knows. Ours is a different case in which an answer exists but is not in the corpus. Their most useful finding for us is that fine-tuning a model to reason more degrades its abstention by roughly 24\%, so a stronger reasoning model should not be assumed to know its limits better.

Wen et al. {[}13{]} give a broader overview of abstention, splitting it into three cases: the query has no answer, the model does not know it, or human values require refusal. Our system addresses the first case. Machcha et al. {[}14{]} introduced MedAbstain and found that simply giving the model an explicit option to abstain does far more than perturbing its input, while making the model larger or the prompt cleverer does almost nothing. We see the same pattern in Section 5.3, where a code-level constraint worked after two prompt-level attempts had failed. The negative-rejection axis of RGB {[}15{]} is an earlier formulation of the same capability. Kalai et al. {[}16{]} make the framing argument: benchmarks reward guessing over admitting uncertainty, so the fix is to reward calibrated abstention explicitly.

\subsection*{2.3 Citation fabrication}\label{citation-fabrication}

A distinct and, through 2026, rapidly growing literature measures citation fidelity in generated scientific text. CiteAudit {[}17{]} benchmarks verification of scientific references with 6,475 real and 2,967 fabricated citations. CiteCheck {[}18{]} performs retrieval-grounded detection of citation hallucinations. Li et al.{[}19{]} introduce the CiteTracer framework, which usefully separates existence-checking ("does this source exist?") from attribution ("does it support the claim?"), with a taxonomy over real, potential, and hallucinated citations. Rao and Callison-Burch {[}20{]} evaluate BibTeX field-level accuracy across 931 papers and report that the DOI is consistently the worst-performing field, which directly motivates the tagged-reference design in Section 3.2.2. A large-scale audit {[}21{]} estimates on the order of 146,000 hallucinated citations in the 2025 preprints and journal literature alone. All of this previous work measures fabrication in generated text, whereas we tie the fabrication of facts to corpus-withholding abstention in the form of a confabulation axis.

\subsection*{2.4 Faithfulness judging}\label{faithfulness-judging}

We score answer faithfulness with MiniCheck {[}7{]}, whose Flan-T5-Large variant is under one billion parameters and reaches GPT-4-comparable accuracy on grounding benchmarks at a fraction of the inference cost. We decide against comparison with a frontier-API judge, since this would mean sending evaluation queries and answers off-premise, which would be in direct contradiction to the design philosophies of the work. We validate the judge against RAGTruth {[}8{]}, a span-annotated hallucination corpus with question-answering, summarisation, and data-to-text splits. Related benchmarks that deserve to be mentioned as context but are not run within our framework include HalluLens {[}25{]}, FaithBench {[}26{]}, and FACTS Grounding {[}28{]}. CSFCube {[}27{]} is a faceted scientific retrieval collection that could be used in other projects, but its computer-science domain does not match our domain that is strongly grounded in biophysics and biochemistry.

\subsection*{2.5 What is new}\label{what-is-new}

Corpus-grounded abstention by itself is, of course, not new. The Framework we propose includes a benchmark that measures both abstention, tested by withholding real source papers from a live retrieval stack while the ingested version of each paper serves as a positive control, as well as confabulation, tested on papers that do not exist and scored by checking emitted DOIs against corpus membership, which requires no judge. The abstention literature covers the first and the citation-fabrication literature covers the second. Pairing them is our novel contribution, and we are not aware of prior work that does so for a paper-search agent.

\textbf{Table 2.1. Coverage of the closest prior work across the four properties our benchmark combines.}

{\small
\begin{longtable}[]{@{}
  >{\raggedright\arraybackslash}p{(\columnwidth - 8\tabcolsep) * \real{0.1900}}
  >{\raggedright\arraybackslash}p{(\columnwidth - 8\tabcolsep) * \real{0.1500}}
  >{\raggedright\arraybackslash}p{(\columnwidth - 8\tabcolsep) * \real{0.1600}}
  >{\raggedright\arraybackslash}p{(\columnwidth - 8\tabcolsep) * \real{0.2900}}
  >{\raggedright\arraybackslash}p{(\columnwidth - 8\tabcolsep) * \real{0.2100}}@{}}
\toprule\noalign{}
\begin{minipage}[b]{\linewidth}\raggedright
\textbf{Work}
\end{minipage} & \begin{minipage}[b]{\linewidth}\raggedright
\textbf{Domain}
\end{minipage} & \begin{minipage}[b]{\linewidth}\raggedright
\textbf{DOI-level ground truth}
\end{minipage} & \begin{minipage}[b]{\linewidth}\raggedright
\textbf{Withholds real documents from a live corpus}
\end{minipage} & \begin{minipage}[b]{\linewidth}\raggedright
\textbf{Confabulation axis}
\end{minipage} \\
\begin{minipage}[b]{\linewidth}\raggedright
KnowOrNot {[}9{]}
\end{minipage} & \begin{minipage}[b]{\linewidth}\raggedright
Government policy
\end{minipage} & \begin{minipage}[b]{\linewidth}\raggedright
No
\end{minipage} & \begin{minipage}[b]{\linewidth}\raggedright
Leave-one-out on a static corpus
\end{minipage} & \begin{minipage}[b]{\linewidth}\raggedright
No
\end{minipage} \\
\begin{minipage}[b]{\linewidth}\raggedright
Sufficient Context {[}10{]}
\end{minipage} & \begin{minipage}[b]{\linewidth}\raggedright
General QA
\end{minipage} & \begin{minipage}[b]{\linewidth}\raggedright
No
\end{minipage} & \begin{minipage}[b]{\linewidth}\raggedright
No
\end{minipage} & \begin{minipage}[b]{\linewidth}\raggedright
No
\end{minipage} \\
\begin{minipage}[b]{\linewidth}\raggedright
RefusalBench {[}11{]}
\end{minipage} & \begin{minipage}[b]{\linewidth}\raggedright
General QA
\end{minipage} & \begin{minipage}[b]{\linewidth}\raggedright
No
\end{minipage} & \begin{minipage}[b]{\linewidth}\raggedright
No (perturbs context)
\end{minipage} & \begin{minipage}[b]{\linewidth}\raggedright
No
\end{minipage} \\
\begin{minipage}[b]{\linewidth}\raggedright
AbstentionBench {[}12{]}
\end{minipage} & \begin{minipage}[b]{\linewidth}\raggedright
General
\end{minipage} & \begin{minipage}[b]{\linewidth}\raggedright
No
\end{minipage} & \begin{minipage}[b]{\linewidth}\raggedright
Not corpus-relative
\end{minipage} & \begin{minipage}[b]{\linewidth}\raggedright
No
\end{minipage} \\
\begin{minipage}[b]{\linewidth}\raggedright
CiteAudit / CiteCheck {[}17, 18{]}
\end{minipage} & \begin{minipage}[b]{\linewidth}\raggedright
Scientific text
\end{minipage} & \begin{minipage}[b]{\linewidth}\raggedright
Partial
\end{minipage} & \begin{minipage}[b]{\linewidth}\raggedright
No
\end{minipage} & \begin{minipage}[b]{\linewidth}\raggedright
Yes
\end{minipage} \\
\begin{minipage}[b]{\linewidth}\raggedright
This work
\end{minipage} & \begin{minipage}[b]{\linewidth}\raggedright
Scientific literature
\end{minipage} & \begin{minipage}[b]{\linewidth}\raggedright
Yes, by construction
\end{minipage} & \begin{minipage}[b]{\linewidth}\raggedright
Yes (paired shadow corpus)
\end{minipage} & \begin{minipage}[b]{\linewidth}\raggedright
Yes
\end{minipage} \\
\midrule\noalign{}
\endhead
\bottomrule\noalign{}
\endlastfoot
\end{longtable}
}

Treating outbound network access as an experimental variable rather than a fixed security policy distinguishes our evaluation from prior agentic-AI security work. Standard practice {[}32{]} and practitioner guidance on air-gapped deployment both recommend denying egress by default and routing traffic through filtering proxies. We instead vary egress between runs and measure its effect on task performance. The distinction is not cosmetic: in Section 5.1 the setting is worth between 0.17 and 0.21 of accuracy on the same questions, and the shadow-corpus test in Section 5.3.2 is only a test of abstention at all when outbound access is disabled.

Our threshold-based certification procedure instantiates a pattern well established in acceptance-test-driven evaluation {[}33{]}, runtime compliance monitoring {[}34{]}, and pre-deployment assurance for enterprise agents {[}35{]}; whether such thresholds reliably predict field behaviour remains an open question {[}36, 37{]}. What we contribute is a concrete implementation for one model, harness, and corpus, shipped inside the system it checks. Our own thresholds are provisional and unvalidated, as stated in Section 4.2.

\section*{3 The Munin Framework}\label{the-munin-framework}

\begin{figure}[tbp]
\centering
\includegraphics[width=\textwidth]{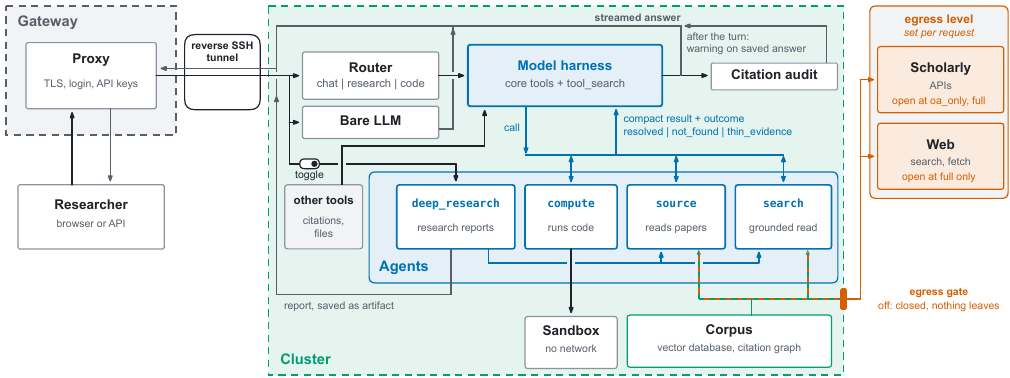}
\caption*{\textbf{Fig. 1 Deployment boundary and one turn through the harness.} All inference, agents, the sandbox, and the corpus live on the cluster\textquotesingle s own machines. The gateway reaches them only through a reverse SSH tunnel initiated from the cluster side, so no inbound port is exposed. When a query arrives, a per-turn router decides to which persona it is passed. The bare LLM is also accessible. Inside the harness, the model invokes agents through core tools and tool\_search. Source and search agents return compact results tagged with an outcome (resolved, not\_found, or thin\_evidence), while keeping the full document text in scope. Deep research is available behind a user-facing toggle and writes its report as a saved artifact. After the turn completes, the citation audit runs separately and attaches a warning to the stored answer if any cited reference cannot be verified. Outbound network access is controlled at the egress gate on a per-request basis: off blocks all external calls, oa\_only permits scholarly APIs, and full additionally opens web search and fetch. If the agents determine that the source for a question is absent from the corpus, they return not\_found. From there, whether to attempt an answer or to decline is the model\textquotesingle s own decision, and a decline is emitted over the same streaming path as any other response (Section 5.3).}\label{fig:architecture}
\end{figure}

\subsection*{3.1 Deployment topology}\label{deployment-topology}

Fig. 1 gives an overview of the framework. The Munin framework comprises a frontend and a backend, deliberately decoupled so that each half can be deployed independently. The backend carries the agent harness, the knowledge base (vector store, citation graph, PDF parser, ingestion pipeline), and the inference endpoint; it runs entirely within the deploying institution. The frontend provides authentication, file upload, and a chat interface, and may be co-located with the backend or hosted externally to expose the system to the public internet. A group with an existing interface can omit the shipped frontend and consume the backend\textquotesingle s OpenAI-compatible API directly.

Every backend service except the retrieval API binds to localhost. The frontend reaches the retrieval API over an outbound reverse tunnel that the backend initiates, so the machine holding the corpus accepts no inbound connections. A chat interface can therefore be exposed on the public internet without opening a port into the research network. Identity arrives at the backend as headers set by the frontend\textquotesingle s forward-authentication layer, so the backend never handles credentials.

\subsection*{3.2 The agent harness}\label{the-agent-harness}

A user turn is routed to one of three profiles before the first model call. The system prompt is composed from a pinned base identity plus a profile-specific fragment. An outer model loop then runs over four named sub-agents (composite tools with isolated interior context) and approximately 45 single-shot tools. Post-turn audits check for phantom URLs, phantom papers, and ungrounded citation claims. Output streams over a resumable server-sent-events channel that can be promoted to a background task if the client disconnects.

Three properties distinguish this architecture from a flat ReAct loop. First, the profile is decided before the first model call rather than by the model mid-turn. Second, sub-agents hold their own context and return compact grounded results, so a 15,000-to-35,000-token paper never enters the outer window. Third, every returned envelope states which provenance guarantee it carries: whether the result was executed or merely parsed, read in full or drawn from an abstract, sourced from the curated corpus or from the open web.

\subsubsection*{3.2.1 Per-turn routing}\label{per-turn-routing}

Routing follows a cheapest-first cascade evaluated before the first model call. Explicit slash commands (/research, /code, /chat) force a specific profile and override both the user\textquotesingle s pin and the fallback vote. The command token is stripped before the query reaches the model. Otherwise, the query is embedded with a small sentence encoder (BGE-base-en-v1.5, 768 dimensions, no query prefix), a lighter encoder than the BGE-large retrieval model, and a distance-weighted k-nearest-neighbour vote is computed over a labelled example set of 244 queries, with the user\textquotesingle s pinned profile injected as a prior. The vote is accepted only if two conditions hold: the winning profile leads the runner-up by at least 0.10 of the total vote mass, and the nearest neighbour is in distribution (cosine similarity of at least 0.45). If either condition fails, the pinned profile applies; if no pin is set, the chat profile is used. Parameters: k = 8, margin threshold 0.10, out-of-distribution threshold 0.45, pin prior weight 0.5. The embedder is injected as a function, so routing is unit-testable offline without loading the encoder.

A harness change is not deployed until it passes the anchor evaluation. It runs a set of 18 hand-written test cases through the full pipeline, including the routing step, and passes a case when the expected tools fire and forbidden ones do not. Because several cases sit legitimately on a profile boundary, the profile the router chose for each case is logged as a diagnostic rather than scored. The anchor set comprises 18 hand-written test queries, disjoint from the router\textquotesingle s example set under exact match (a collision check runs whenever the example set is rebuilt). The trajectory pass rate across gating runs exceeded the 0.95 threshold. On the 10 queries with a known expected profile, the router selects the correct one in 7 cases.

This design replaces an earlier mechanism in which one persona could hand a turn to another mid-conversation. A query classified as chat but requiring a research answer now routes to the research profile on the same turn, eliminating the need for an intra-turn delegation step. Removing the per-persona tool allowlists made delegation unnecessary for access purposes as well, since every tool is reachable through a core set plus tool search regardless of profile. Guards are therefore per-tool rather than per-profile and are checked in code before invocation: the sandbox runs with its network namespace disabled and a fixed execution ceiling, every network-reaching tool honours the per-request egress level, web fetches are limited to URLs that appeared in a prior result, and the loop carries a cumulative tool-call cap and a turn budget.

\subsubsection*{3.2.2 source: full-text reading}\label{source-full-text-reading}

The source agent answers a question against one or more specific documents and exposes four modes: summary, question-answering, extraction, and comparison. Each call runs the same three steps: resolve the reference to a document, extract its full text, and answer using the entire text in a single language-model call. Only documents that exceed the context window are chunked instead. In Appendix B.2 we investigate the reading path more and find that replacing a summarise-then-reason predecessor moved answer accuracy from 0.497 to 0.864.

A reference may be given as a DOI, an arXiv identifier, an internal identifier, a URL, or a title with optional authors and year. Accepting all five forms addresses a specific failure mode that a stricter interface would introduce. If the interface requires a DOI, the outer model must supply one for every document it wants to read, and when it does not possess the correct DOI it produces a plausible one instead. DOIs invite this more than any other field: their structured, publisher-prefixed format means that a fabricated DOI is indistinguishable from a real one. Rao and Callison-Burch {[}20{]} measured field-level accuracy across 931 papers and found the DOI to be the least accurate field in generated bibliographies.

The worst case is not the DOI that fails to resolve. It is the DOI that resolves to a real but incorrect publication source. The agent reads that document, quotes it accurately, and returns a result whose grounding is intact in every respect except that the source is not the one the user asked about. No check in the system alarms, because nothing has gone wrong except the identity of the source. A human scanning the trace would catch a wrong title immediately, but cannot recognise a wrong DOI from its format alone.

Accepting titles removes the pressure to fabricate, and tagging the reference type makes the remaining cases auditable. A DOI-tagged reference that appears in neither the user\textquotesingle s input nor any earlier tool output is a candidate for model fabrication. Logging each such event converts what would otherwise be a suspicion into a countable signal, and in aggregate a measurable rate.

A single boolean (found / not-found) collapses distinct failure modes that require different responses from the user or operator (see table 7): the document has been read but does not contain the answer; it has been located but could not be parsed; the reference matched several documents; it exists outside the active scope; or it is not obtainable. Only the first case is an informative negative; the remaining four are system failures that a user or operator can address.

\textbf{Table 3.1. Outcome values returned by the reading agent.}

{\small
\begin{longtable}[]{@{}
  >{\raggedright\arraybackslash}p{(\columnwidth - 4\tabcolsep) * \real{0.2200}}
  >{\raggedright\arraybackslash}p{(\columnwidth - 4\tabcolsep) * \real{0.4200}}
  >{\raggedright\arraybackslash}p{(\columnwidth - 4\tabcolsep) * \real{0.3600}}@{}}
\toprule\noalign{}
\begin{minipage}[b]{\linewidth}\raggedright
\textbf{Outcome}
\end{minipage} & \begin{minipage}[b]{\linewidth}\raggedright
\textbf{Meaning}
\end{minipage} & \begin{minipage}[b]{\linewidth}\raggedright
\textbf{Caller contract}
\end{minipage} \\
\begin{minipage}[b]{\linewidth}\raggedright
not\_found
\end{minipage} & \begin{minipage}[b]{\linewidth}\raggedright
Document parsed successfully; answer not present
\end{minipage} & \begin{minipage}[b]{\linewidth}\raggedright
Treated as a definitive negative
\end{minipage} \\
\begin{minipage}[b]{\linewidth}\raggedright
extraction\_failed
\end{minipage} & \begin{minipage}[b]{\linewidth}\raggedright
Reference resolved but document unreadable (scanned PDF, parser error)
\end{minipage} & \begin{minipage}[b]{\linewidth}\raggedright
Logged for re-ingestion
\end{minipage} \\
\begin{minipage}[b]{\linewidth}\raggedright
ambiguous(candidates)
\end{minipage} & \begin{minipage}[b]{\linewidth}\raggedright
Fuzzy reference matched multiple documents
\end{minipage} & \begin{minipage}[b]{\linewidth}\raggedright
Requires explicit disambiguation; top-hit auto-selection is prohibited
\end{minipage} \\
\begin{minipage}[b]{\linewidth}\raggedright
out\_of\_scope
\end{minipage} & \begin{minipage}[b]{\linewidth}\raggedright
Document exists in the full corpus but outside the active sub-corpus
\end{minipage} & \begin{minipage}[b]{\linewidth}\raggedright
User informed that widening scope would resolve it
\end{minipage} \\
\begin{minipage}[b]{\linewidth}\raggedright
resolution miss
\end{minipage} & \begin{minipage}[b]{\linewidth}\raggedright
Not in the corpus and not downloadable
\end{minipage} & \begin{minipage}[b]{\linewidth}\raggedright
Logged as a corpus gap
\end{minipage} \\
\midrule\noalign{}
\endhead
\bottomrule\noalign{}
\endlastfoot
\end{longtable}
}

The abstention benchmark in Section 4.6 requires the system to report which type of failure occurred, because calibrated refusal, confabulation, and mechanical failure demand different responses from a user. The outcome flag handles the first separation: \emph{not\_found} signals an informative negative, while the other four outcomes signal addressable system failures. Confabulation is detected separately by the oracle in Appendix B.2, which compares the agent\textquotesingle s answer against the source and flags cases where no supporting sentence exists. Neither mechanism substitutes for the other: the flag is blind to confabulation (which produces a positive answer, not a negative outcome), and the oracle does not diagnose which mechanical step failed when parsing goes wrong.

Beyond the failure cases, two properties constrain the agent\textquotesingle s positive answers. First, the quote-or-abstain contract requires the agent to cite the exact sentence supporting its answer, or to return \emph{not\_found}. This makes every positive answer verifiable against the source text. The contract does not protect against misreading: the agent can quote a real sentence and interpret it incorrectly. The oracle in Appendix B.2 observed this in approximately 9\% of positive answers. Second, in extraction mode the agent returns a handle rather than the extracted rows. Because the outer model never receives the raw values, it cannot reproduce them, regardless of how it is prompted. Five preview rows allow it to describe the data shape to the user without making reconstruction possible from context alone.

\subsubsection*{3.2.3 search: tiered evidence retrieval}\label{search-tiered-evidence-retrieval}

This agent puts local corpus search, scholarly APIs, and web search behind a single call. Which tiers it actually reaches is set by the egress control rather than by the model: with egress off, it searches only the local corpus; open-access mode adds the scholarly APIs and full mode adds the web. In Sections 4.2 and 5.1 we treat this control as an experimental variable.

Every result carries its source type, which is what makes ranking across tiers possible. Trust is not uniform across a corpus paper, an open-access paper, and a web page, so a single similarity score cannot order them: a Nature paper and a search-optimised listicle are not comparable on embedding distance. The ranker therefore allocates fixed slots per tier, three for web and four for open access, rather than computing one global score. Without quotas, the web tier dominates on volume and shallow lexical overlap.

External slots must additionally clear a relevance floor, set at 0.62. This value is specific to the encoder and was calibrated over 30 real retrieved candidates, where BGE cosine similarities occupy a compressed band of roughly 0.57 to 0.77. It is not a general threshold for "similar enough" and does not transfer: a group deploying with a different encoder or in a different domain must recalibrate it. The default ships as a configuration parameter (Section 3.4).

Deduplication compares an alias set of DOIs, arXiv identifiers, and normalised titles rather than any single field. A preprint and its published version constitute one work with two DOIs; without alias matching, a comparison call treats the pair as two distinct documents.

The coverage note reports the number of results within the active scope alongside the total across the full corpus, and appears whenever a sub-corpus filter is applied. Its purpose is to prevent a filter effect from being read as a corpus gap: a user told that nothing was found will conclude the literature does not contain the topic, when in fact the material was excluded by their own scope. The unscoped query is the same query, so the second count is nearly free. When strong scholarly hits number fewer than three, the thin-evidence flag marks the result set as sparse rather than padding it with weak matches to reach a fixed count.

\subsubsection*{3.2.4 compute: sandboxed code generation}\label{compute-sandboxed-code-generation}

The language a request is written in determines how the resulting code is checked, and that mapping is fixed in the harness rather than chosen by the model. Python is executed in a sandbox, while other languages receive the strongest static check available for them: compilation for C++, parsing with linting for R. Asking the model whether a given task needs sandboxing would insert a stochastic judgment into a pipeline built for auditability, a judgment the model has an incentive to answer in the direction that costs it least.

The guarantees that the code returned by the agent is correct differ by language. C++ compilation performs full type checking, so a syntax-only build catches what R linting cannot. Python execution catches runtime errors that no static check on C++ would surface. Every code result includes a verification label: executed, compiled, or linted. Without it, a user receiving R code has no signal that it was never run, and may read it as carrying the same guarantee as executed Python. For plots the static rungs are worth very little: a script that selects the wrong column, mislabels an axis, or plots against the index passes linting cleanly. Only execution distinguishes those cases.

The sandbox is active from the first import, not only during the test phase, because importing a module executes its top-level code. The environment is a single pinned scientific image with no package installation, no network namespace, a read-only filesystem with one scratch directory, and caps on memory, processes, and wall-clock time. Thread counts and CPU quota are pinned so that the agent can be throttled to one or two cores when the cluster is inference-only, rather than being shut down entirely.

The repair loop is bounded at three attempts and halts immediately on a repeated error signature, hashed from exception type and final stack frame, since an identical recurring error indicates the model is stuck rather than converging. The loop also never sees model-authored tests, since an agent that writes its own tests will likely converge on passing them by weakening them, so tests come from the specification or from a separate generation step.

Results are delivered as a bundle of script, data file, and figure, with the script reading its data from its own directory, which is the only arrangement in which the delivered code is both byte-identical to what ran and runnable standalone. Whether a value may appear as a literal is governed by origin rather than syntax: data the user typed or uploaded, or data a script generates, may be inlined, while data drawn from a document arrives only as a handle. Banning literals outright is unenforceable, because syntax cannot distinguish a parameter from a datum.

Executing the code verifies that it runs without error; it does not verify that the result is correct. A script can execute cleanly and still select the wrong variable, apply the wrong formula, or misread a unit. The final check is the user\textquotesingle s, and returning the rendered figure alongside the code makes that check more tractable than a bare numeric output: a mislabeled axis or an inverted trend is visible at a glance, whereas a wrong scalar value embedded in a text response is not.

\subsubsection*{3.2.5 deep\_research: planned multi-source synthesis}\label{deep_research-planned-multi-source-synthesis}

The deep research agent runs in-process against the live inference server. A run lasts several minutes, so it must not block a chat turn. The agent decomposes the question into sub-questions and works through them adaptively. Information from one paper can add new sub-questions or change which one is pursued next.

The agent first develops a research plan in the form of a data structure rather than a chain of reasoning. It holds a list of sub-questions, each with an id, a status of open, resolved, or unresolvable, notes, and evidence references. The loop selects an open sub-question, searches, screens the results, reads the survivors, writes notes under that sub-question\textquotesingle s id, updates the status, and may add new sub-questions. An earlier design interleaved reasoning and tool calls in a single loop, without an explicit plan. In practice the model collapsed into a repetitive sequence of tool calls at greater length and cost, with no visible planning step between calls.

Three properties follow from the plan being an object. Termination is defined by every sub-question being resolved or marked unresolvable, with budget exhaustion as a floor rather than the normal stopping condition. Unresolved sub-questions can be reported to the user instead of quietly omitted. The plan together with its notes constitutes the complete job state, so a run interrupted at minute eighteen resumes rather than restarts.

The pipeline retrieves roughly 100 candidates, selects 25 to 30 for full reading, then produces notes, an outline, and sections. The entire run uses about 35 model calls. These limits are fixed parameters rather than agent decisions, making cost predictable in advance. Screening is inexpensive because it is batched, since 100 abstracts at roughly 200 tokens each fit in a single call.

The screener\textquotesingle s two error types are not symmetric. A false positive costs one unnecessary read and is visible in the trace. A false negative is silent and unrecoverable: the paper is in the corpus, the search retrieves it, the screener discards it, and the answer never appears in the final document. To the user, this is indistinguishable from the corpus not containing the material. The screener is therefore tuned for recall: it errs toward keeping, and it logs every discarded candidate with its reason. Recording reading depth per source gives the screener a third option: retain a paper at abstract depth instead of choosing between a full read and a discard.

Citations found within the papers read in full are pursued to depth one only and are routed back through the screener rather than read directly. The depth limit is a tuning parameter, set to one because it bounds the search space and keeps per-run cost predictable., The (likely) marginal yield of a second hop has not been measured by us, but this depth limit can be modified in other deployments.

Each note is a structured record with fixed fields: claim, value, unit, supporting quote, reference, and sub-question id. The schema exists to make cross-source conflicts detectable. When multiple papers address the same question their values often disagree, and the synthesis must surface that disagreement rather than silently averaging it. Free-text notes would lose it: the synthesiser drafts section by section, so two papers with contradictory values may never share a context window. With structured records, all notes under a given sub-question id are grouped and their values compared mechanically, and a discrepancy is flagged without an additional model call. The synthesiser then reports both values, resolves the conflict, or flags the uncertainty to the user.

Grounding passes through three lossy hops, from quote to note to section prose to finished document. The synthesiser therefore drafts from the original quotes rather than from paraphrases of paraphrases. Each citation also records the reading depth of its source: full text, abstract, or snippet. A document citing 100 sources, of which 70 were accessed only via their abstracts, should say so per citation; without that, "synthesised from 100 sources" overstates the work and gives the reader no way to judge which citations to weigh.

Sections retrieve their notes by sub-question id, not by similarity. The plan already records which notes belong to which question, so selection is a structural lookup with no vector search and no retrieval-miss failure mode. This imposes a constraint, as the synthesiser may order and merge sections but cannot introduce one with no sub-question behind it, because no notes would exist to draft it from. However, this means that every section maps cleanly to a question the plan posed.

\subsubsection*{3.2.6 Guardrails and audits}\label{guardrails-and-audits}

\textbf{Table 3.2. Mechanisms that constrain or check the outer loop.}

{\small
\begin{longtable}[]{@{}
  >{\raggedright\arraybackslash}p{(\columnwidth - 2\tabcolsep) * \real{0.2800}}
  >{\raggedright\arraybackslash}p{(\columnwidth - 2\tabcolsep) * \real{0.7200}}@{}}
\toprule\noalign{}
\begin{minipage}[b]{\linewidth}\raggedright
\textbf{Mechanism}
\end{minipage} & \begin{minipage}[b]{\linewidth}\raggedright
\textbf{What it does}
\end{minipage} \\
\begin{minipage}[b]{\linewidth}\raggedright
Tool-call cap (30 per answer)
\end{minipage} & \begin{minipage}[b]{\linewidth}\raggedright
Hard code limit; on reaching it, control routes to the existing wrap-up synthesis.
\end{minipage} \\
\begin{minipage}[b]{\linewidth}\raggedright
Phantom-URL audit
\end{minipage} & \begin{minipage}[b]{\linewidth}\raggedright
Flags URLs in the answer that appear in no tool result
\end{minipage} \\
\begin{minipage}[b]{\linewidth}\raggedright
Phantom-paper audit
\end{minipage} & \begin{minipage}[b]{\linewidth}\raggedright
Flags cited papers that appear in no tool result
\end{minipage} \\
\begin{minipage}[b]{\linewidth}\raggedright
Citation-claim grounding audit
\end{minipage} & \begin{minipage}[b]{\linewidth}\raggedright
Flags any "X et al." whose surname appears in no tool result from that turn. Matching is word-boundary, not substring, because a substring match would miss a surname embedded inside a longer token in the payload.
\end{minipage} \\
\begin{minipage}[b]{\linewidth}\raggedright
Identifier resolution
\end{minipage} & \begin{minipage}[b]{\linewidth}\raggedright
DOI, PMID, PMCID, and arXiv identifiers resolve through corpus, scholarly APIs, and registries. No language model is in this path; the resolver cannot produce a fabricated identifier.
\end{minipage} \\
\begin{minipage}[b]{\linewidth}\raggedright
Explicit null authors
\end{minipage} & \begin{minipage}[b]{\linewidth}\raggedright
Where metadata cannot be obtained, the result carries an explicit null and a flag. A missing field reads to a model as "not applicable"; an explicit null reads as "unknown, do not guess."
\end{minipage} \\
\begin{minipage}[b]{\linewidth}\raggedright
Transport retry
\end{minipage} & \begin{minipage}[b]{\linewidth}\raggedright
All inference calls retry on 5xx, 429, and pre-first-byte stream drops with exponential backoff, honouring Retry-After.
\end{minipage} \\
\midrule\noalign{}
\endhead
\bottomrule\noalign{}
\endlastfoot
\end{longtable}
}

Rather than instructing the model to behave well, the harness either makes the failure structurally impossible or detects it after the fact and flags it before it reaches the user. Structural impossibility means the code path cannot produce the bad output at all; post-hoc detection means the bad output can occur but is caught before delivery. In extraction mode, for instance, the agent returns an opaque reference to the data rather than the data itself, so the outer model cannot transcribe values it never received. The mechanisms in Table 3.2 span both: the tool-call cap and identifier resolution belong to the first class, while the phantom-URL, phantom-paper, and citation-claim audits belong to the second.

The tool-call cap is a code-level limit rather than a prompt instruction, because two prompt-level interventions against over-tooling failed (Appendix B.4). Identifier resolution guarantees that any identifier reaching the corpus is real. It does not catch the case where the outer model invents an identifier before calling the resolver; that is handled by the tagged-reference check in Section 3.2.2, which flags any identifier absent from the user\textquotesingle s input and all prior tool outputs.

Every agent also emits a trace recording its resolved sources, model calls, decisions, and any abstention or failure reason. The evaluation suite reads these traces rather than the final answers alone, so that Section 5.3 can report why an answer was withheld, not merely that it was.

\subsection*{3.3 The knowledge base}\label{the-knowledge-base}

The corpus is the group\textquotesingle s collective reading rather than a slice of a public database. Members upload the papers they are currently working with, and the index grows along the group\textquotesingle s citation neighbourhood rather than along a topic taxonomy specified in advance. Our deployment spans chemistry, biophysics, and membrane biology, but the framework itself is domain independent. A different group uploading a few thousand papers produces a corpus shaped by their own reading. At present the corpus holds 3,922 records from direct contributor uploads. A citation-graph crawler, seeded from those records, resolves references of each ingested paper through institutional access and open-access sources, growing the total to 68,500.

In our deployment the corpus is consortium-wide, and sub-corpora selected by slash command are a relevance filter rather than a hard boundary. Sub-corpus membership is a label set rather than a partition, applied at ingestion, because a paper can be relevant to several groups at once. Ingestion effort therefore compounds across the consortium: a paper one group adds is immediately available to answer another group\textquotesingle s question, and a query that would find nothing against a single laboratory\textquotesingle s reading list can find its answer in a neighbouring group\textquotesingle s.

Because there is no access boundary, the summary cache can key on the document identifier alone. Under a per-group corpus, the same cache would return a summary of a paper the requester is not authorised to access. Collective visibility is what makes an aggressive cache safe here; a deployment requiring per-group isolation would have to forgo caching altogether.

A separate index over members\textquotesingle{} own uploaded documents runs alongside the paper corpus so that notes, preprints, and internal reports are reachable without being mixed into the literature index. It uses its own collection and its own embedding model.

\subsubsection*{3.3.1 Ingest and the identity guard}\label{ingest-and-the-identity-guard}

Each PDF is parsed into structured XML and enriched against a metadata registry. The resulting record is embedded and written to the vector store, the citation graph, and a state database. Embeddings are built from title and abstract only, so correcting an author field later does not require re-embedding.

Enrichment is the step most prone to silent error. The parser sometimes extracts a reference\textquotesingle s DOI rather than the paper\textquotesingle s own, so a paper is keyed under a cited work\textquotesingle s identifier and then enriched with that work\textquotesingle s metadata. The result is one record holding two papers\textquotesingle{} metadata spliced together. The pipeline compares the registry title against the parser-extracted title using Jaccard similarity over token sets and rejects the enrichment when the score falls below a fixed threshold. On rejection it retains the parser-extracted metadata, discards the registry values, and stores the conflicting title in the record so that a maintenance pass can locate and correct it.

\subsubsection*{3.3.2 Continuous quality control}\label{continuous-quality-control}

The corpus is maintained by a scheduled detection service that runs every fifteen minutes. It flags metadata mismatches, low-quality or short records, unverifiable metadata, and orphaned entries, and moves flagged records to quarantine. An operator triages the quarantine weekly. Two nightly jobs handle bookkeeping: one backfills group attribution for records whose uploader joined the contributor allowlist after ingestion, and the other rebuilds the two-dimensional embedding map and its topical clusters.

Quarantine is reversible. A quarantined paper leaves the searchable corpus but its PDF is moved rather than deleted, and its citation-graph node is retained so that removing a paper does not fragment the citation graph. Deleting a record is a separate operator command.

The nightly map clusters the corpus by embedding similarity and serves as both a user-facing browsing feature and a corpus diagnostic: a sudden shift in cluster structure is visible in the map before it shows up in query results. Clustering requires L2-normalised vectors. The encoder\textquotesingle s outputs are not guaranteed to be unit-norm; without normalisation, Euclidean distances become dominated by vector magnitude rather than direction, and the density-clustering step collapses to a single large cluster. Points that fall below the cluster-density threshold are labelled unclustered rather than assigned to the nearest cluster. Forcing them in would inflate every cluster and hide noise. Cluster labels are generated from the ten titles nearest each centroid and assigned to all members of that cluster. Labelling therefore requires one model call per cluster rather than per paper.

\subsubsection*{3.3.3 Corpus quality is a measurement concern}\label{corpus-quality-is-a-measurement-concern}

Two standing audits of the reference deployment are reported in Appendix E, because their findings changed how we read the evaluation. A title audit flagged 22.4\% of contributor-upload records as having incorrect titles; manual review showed 30 to 40\% of the flags to be detector errors. An author-metadata repair moved records from 80.3\% to 98.7\% clean. During the repair we found that a major scholarly API returns wrong authors for 3 of 16 validation records and excluded it as a backfill source.

In a private-corpus retrieval system, a substantial fraction of apparent model failures are corpus failures. The two cannot be distinguished without knowing the correct metadata at the time records are ingested and running continuous audits thereafter. This is why a framework of this kind must ship corpus tooling rather than assume a clean corpus.

\subsection*{3.4 Configuration surface}\label{configuration-surface}

Porting the framework to a different corpus or model requires changing five settings (Table 3.3).

\textbf{Table 3.3. Settings requiring deliberate attention.}

{\small
\begin{longtable}[]{@{}
  >{\raggedright\arraybackslash}p{(\columnwidth - 2\tabcolsep) * \real{0.2200}}
  >{\raggedright\arraybackslash}p{(\columnwidth - 2\tabcolsep) * \real{0.7800}}@{}}
\toprule\noalign{}
\begin{minipage}[b]{\linewidth}\raggedright
\textbf{Setting}
\end{minipage} & \begin{minipage}[b]{\linewidth}\raggedright
\textbf{Why it matters}
\end{minipage} \\
\begin{minipage}[b]{\linewidth}\raggedright
Encoder / collection pairing
\end{minipage} & \begin{minipage}[b]{\linewidth}\raggedright
Pairing a new encoder with a collection written by a previous one returns empty or incorrect results rather than raising an error. The pairing is asserted at three points: a startup check comparing encoder width against the collection\textquotesingle s vector size, a deploy-time re-assertion against the live status endpoint, and code defaults that match production. A mismatch fails the boot.
\end{minipage} \\
\begin{minipage}[b]{\linewidth}\raggedright
Query-side embedding prefix
\end{minipage} & \begin{minipage}[b]{\linewidth}\raggedright
Applied to queries only, with documents embedded raw. Encoder-dependent, and empty for some encoders. Omitting it materially reduces recall.
\end{minipage} \\
\begin{minipage}[b]{\linewidth}\raggedright
Served context length
\end{minipage} & \begin{minipage}[b]{\linewidth}\raggedright
Pinned in both the inference server configuration and the backend token budget, and exported as a shared environment variable before the retrieval container is recreated so the two values cannot drift apart.
\end{minipage} \\
\begin{minipage}[b]{\linewidth}\raggedright
Serving concurrency ceiling
\end{minipage} & \begin{minipage}[b]{\linewidth}\raggedright
Exceeding the inference server\textquotesingle s concurrent-sequence limit degrades answer quality silently rather than erroring. Evaluation must run at or below it.
\end{minipage} \\
\begin{minipage}[b]{\linewidth}\raggedright
External-slot relevance floor
\end{minipage} & \begin{minipage}[b]{\linewidth}\raggedright
Calibrated to one encoder over 30 real candidates within an observed similarity band of 0.57 to 0.77. Not a general-purpose threshold: a group using a different encoder or working in a different domain must recalibrate it.
\end{minipage} \\
\midrule\noalign{}
\endhead
\bottomrule\noalign{}
\endlastfoot
\end{longtable}
}

\subsection*{3.5 Deployment boundary}\label{deployment-boundary}

Inference, the corpus, the vector store, the citation graph, the code sandbox, and the faithfulness judge all run inside the deploying institution. Two components necessarily leave the institution, and in both cases placing them externally is the deploying group\textquotesingle s decision rather than a requirement imposed by the framework. In our reference deployment the frontend runs on rented infrastructure, so chat messages pass through a third-party host; backend-only deployment is offered to avoid this. When web search is enabled, query text is sent to a commercial search API.

\section*{4 Experimental Setup}\label{experimental-setup}

\subsection*{4.1 The reference deployment}\label{the-reference-deployment}

As a reference deployment we run Munin on a cluster head node with two NVIDIA RTX 5090 cards (32 GB VRAM each, Blackwell, SM 120a) under CUDA 13.0.2. Job accounting is not deployed, which is why cost in Section 5.6 is reported as tokens plus wall-clock rather than GPU-seconds. The frontend runs on a rented ARM virtual server.

\subsection*{4.2 Backbones, arms and protocol}\label{backbones-arms-and-protocol}

Every result in Section 5 that involves a language model is reported on the three backbones listed in Table 4.2. The two Qwen models are both served at the same context window and concurrency, so that a change of model and a change of context size are not entangled in the measurements. Qwen3.6-35B-A3B is a mixture-of-experts model with 3B active parameters; it served production traffic until August 2026 and is retained here as an evaluation-only instance. Qwen3.8-27B is a dense model and the current production backbone, which makes it the headline configuration for most of the chapter. gpt-oss-20b comes from a different developer and is a 21B mixture-of-experts model with 3.6B active parameters, served natively in its MXFP4 quantisation on a single GPU. Including it changes the developer, the architecture family, and the quantisation format simultaneously, which means its smaller harness effect cannot be attributed to any one of those three factors.

\textbf{Table 4.2 Backbones and serving configurations.}

{\small
\begin{longtable}[]{@{}
  >{\raggedright\arraybackslash}p{(\columnwidth - 6\tabcolsep) * \real{0.2500}}
  >{\raggedright\arraybackslash}p{(\columnwidth - 6\tabcolsep) * \real{0.2500}}
  >{\raggedright\arraybackslash}p{(\columnwidth - 6\tabcolsep) * \real{0.2500}}
  >{\raggedright\arraybackslash}p{(\columnwidth - 6\tabcolsep) * \real{0.2500}}@{}}
\toprule\noalign{}
\begin{minipage}[b]{\linewidth}\raggedright
\end{minipage} & \begin{minipage}[b]{\linewidth}\raggedright
\textbf{Qwen3.6-35B-A3B}
\end{minipage} & \begin{minipage}[b]{\linewidth}\raggedright
\textbf{Qwen3.8-27B}
\end{minipage} & \begin{minipage}[b]{\linewidth}\raggedright
\textbf{gpt-oss-20b}
\end{minipage} \\
\begin{minipage}[b]{\linewidth}\raggedright
Developer
\end{minipage} & \begin{minipage}[b]{\linewidth}\raggedright
Alibaba
\end{minipage} & \begin{minipage}[b]{\linewidth}\raggedright
Alibaba
\end{minipage} & \begin{minipage}[b]{\linewidth}\raggedright
OpenAI
\end{minipage} \\
\begin{minipage}[b]{\linewidth}\raggedright
Architecture
\end{minipage} & \begin{minipage}[b]{\linewidth}\raggedright
MoE, 35B total / 3B active
\end{minipage} & \begin{minipage}[b]{\linewidth}\raggedright
dense 27B
\end{minipage} & \begin{minipage}[b]{\linewidth}\raggedright
MoE, 21B total / 3.6B active
\end{minipage} \\
\begin{minipage}[b]{\linewidth}\raggedright
Quantisation
\end{minipage} & \begin{minipage}[b]{\linewidth}\raggedright
AWQ 4-bit, fp8 KV cache
\end{minipage} & \begin{minipage}[b]{\linewidth}\raggedright
AWQ INT4, fp8 KV cache
\end{minipage} & \begin{minipage}[b]{\linewidth}\raggedright
native MXFP4
\end{minipage} \\
\begin{minipage}[b]{\linewidth}\raggedright
Serving
\end{minipage} & \begin{minipage}[b]{\linewidth}\raggedright
one card, 65,536-token window
\end{minipage} & \begin{minipage}[b]{\linewidth}\raggedright
two cards, tensor-parallel, 65,536-token window
\end{minipage} & \begin{minipage}[b]{\linewidth}\raggedright
one card (24 GB at this window)
\end{minipage} \\
\begin{minipage}[b]{\linewidth}\raggedright
Role
\end{minipage} & \begin{minipage}[b]{\linewidth}\raggedright
production until 2026-08-25; eval-only for these runs
\end{minipage} & \begin{minipage}[b]{\linewidth}\raggedright
production; headline
\end{minipage} & \begin{minipage}[b]{\linewidth}\raggedright
third backbone; eval-only
\end{minipage} \\
\midrule\noalign{}
\endhead
\bottomrule\noalign{}
\endlastfoot
\end{longtable}
}

The harness ablation compares three arms on the same 199 LitQA2 questions {[}1{]} whose source papers are present in the corpus. Questions whose source is absent measure abstention rather than answering and are reserved for the tests in Section 4.6. The bare arm receives only the question and answers from the model\textquotesingle s weights alone. The naive-RAG arm prepends the top-5 documents from the production vector store to the prompt and issues a single completion. The agentic arm is the full harness as described in Section 3. All three arms run at concurrency 1 with a 16,384-token output budget and a 900-second deadline.

Outbound network access is set per request at three levels: off, open-access, and full. The agentic arm is run twice, once at full egress and once with outbound access disabled (the corpus-only arm), so that the total harness value can be decomposed into what the local corpus contributes and what the external tiers add on top. The bare and RAG arms bypass the harness entirely, using frozen prompts and the same token budget on every backbone, which makes them controlled comparisons across models. The agentic arms on the two Qwen backbones were captured one day of harness development apart. The bare and RAG prompts were deliberately not adapted per backbone; doing so would hide the cost that a model trained to expect tools pays when those tools are unavailable, and for that reason Section 5.1 reports precision of attempted answers alongside accuracy. Bare and RAG sample at temperature 0.7 while the agentic arm uses the research profile\textquotesingle s settings; this asymmetry is present in every run and is listed among the limitations in Section 6.2.

Within a backbone, all comparisons are question-paired. Confidence intervals are 95\% percentile bootstraps over 1,000 resamples with a fixed seed (2,000 for the abstention tests in Section 5.3). We measured the run-to-run noise directly: a same-day recapture of the Qwen3.8 agentic arm differs from the headline figure by -0.005 {[}-0.050, +0.040{]}, and two identical bare arms run a day apart differ by 0.035. Differences below approximately three points are therefore treated as within the measurement floor. The same captures that produce the accuracy tables also feed the faithfulness scores and the tool-use telemetry, so every table in Section 5 describes the identical set of runs. Each run is committed as a scorecard carrying its provenance, egress level, sampling configuration, and budget; the evaluation suite, the comparison command, and the certification thresholds are all documented in the repository.

\subsection*{4.3 Retrieval datasets}\label{retrieval-datasets}

\textbf{Table 4.3. Retrieval and answering datasets..}

{\small
\begin{longtable}[]{@{}
  >{\raggedright\arraybackslash}p{(\columnwidth - 6\tabcolsep) * \real{0.2200}}
  >{\raggedright\arraybackslash}p{(\columnwidth - 6\tabcolsep) * \real{0.2200}}
  >{\raggedright\arraybackslash}p{(\columnwidth - 6\tabcolsep) * \real{0.3600}}
  >{\raggedright\arraybackslash}p{(\columnwidth - 6\tabcolsep) * \real{0.2000}}@{}}
\toprule\noalign{}
\begin{minipage}[b]{\linewidth}\raggedright
\textbf{Dataset}
\end{minipage} & \begin{minipage}[b]{\linewidth}\raggedright
\textbf{Size}
\end{minipage} & \begin{minipage}[b]{\linewidth}\raggedright
\textbf{What it tests}
\end{minipage} & \begin{minipage}[b]{\linewidth}\raggedright
\textbf{Domain}
\end{minipage} \\
\begin{minipage}[b]{\linewidth}\raggedright
BEIR / SciFact {[}5, 6{]}
\end{minipage} & \begin{minipage}[b]{\linewidth}\raggedright
300 queries, 5,183 docs
\end{minipage} & \begin{minipage}[b]{\linewidth}\raggedright
External validity anchor; claim verification
\end{minipage} & \begin{minipage}[b]{\linewidth}\raggedright
Biomedical
\end{minipage} \\
\begin{minipage}[b]{\linewidth}\raggedright
LitSearch {[}4{]}
\end{minipage} & \begin{minipage}[b]{\linewidth}\raggedright
597 queries, 64,183 docs
\end{minipage} & \begin{minipage}[b]{\linewidth}\raggedright
The closest public analogue to actual usage
\end{minipage} & \begin{minipage}[b]{\linewidth}\raggedright
ML / NLP
\end{minipage} \\
\begin{minipage}[b]{\linewidth}\raggedright
LitQA2 retrieval {[}1{]}
\end{minipage} & \begin{minipage}[b]{\linewidth}\raggedright
199 questions
\end{minipage} & \begin{minipage}[b]{\linewidth}\raggedright
Retrieval and end-to-end answering on in-corpus questions
\end{minipage} & \begin{minipage}[b]{\linewidth}\raggedright
Biology, in our corpus
\end{minipage} \\
\begin{minipage}[b]{\linewidth}\raggedright
Encoder bake-off pool
\end{minipage} & \begin{minipage}[b]{\linewidth}\raggedright
190 sources + 5,000 distractors
\end{minipage} & \begin{minipage}[b]{\linewidth}\raggedright
Encoder comparison on pre-embedded pools
\end{minipage} & \begin{minipage}[b]{\linewidth}\raggedright
Our domain
\end{minipage} \\
\midrule\noalign{}
\endhead
\bottomrule\noalign{}
\endlastfoot
\end{longtable}
}

The evaluation suite includes a validation gate whose purpose is to confirm that the retrieval pipeline correctly ranks documents before any model-specific results are interpreted. The specified criterion for the reference deployment was that the production encoder should achieve at least 0.5 nDCG@10 (a ranking-quality score in {[}0, 1{]} that rewards placing relevant documents high in the result list) on the SciFact collection. When the gate ran, before the migration to BGE-large, the encoder then in production, SPECTER, scored 0.479 as deployed and 0.494 with its canonical {[}SEP{]} input format, both just below the threshold. The production encoder, BGE-large, was evaluated on LitSearch instead (Section 5.5).

Rather than adjusting the threshold or setting the result aside, we tested whether the harness itself was producing reliable rankings by reproducing the published BM25 baseline on the same collection. The measured score was 0.652, compared with the published value of 0.665 {[}5{]}. A difference of 0.013 is consistent with known variations in tokenisation and implementation detail between the original experimental setup and ours. This reproduction shows that when a well-characterised classical retriever is run through the same pipeline, it recovers the expected ranking quality on this collection, which means the pipeline itself is not distorting the signal.

The practical consequence of the gate passing in this manner is that subsequent scores can be attributed to the model under test. If the encoder scores below threshold on a future benchmark run, the operator can trust that the cause is the encoder\textquotesingle s performance on that domain, not a defect in the retrieval infrastructure. Validating the measurement apparatus is the primary function of a retrieval gate: a score produced by a faulty pipeline tells the operator nothing about the retriever, whereas a score produced by a validated pipeline isolates the variable of interest.

\subsection*{4.4 Metrics and statistics}\label{metrics-and-statistics}

We reimplement the retrieval metrics inside the harness and verify the implementation against a standard information retrieval (IR) evaluation library, confirming that the two agree bit-for-bit. Reimplementing rather than linking against the library keeps the harness free of heavy dependencies and makes every metric convention explicit in the source code rather than implicit inside a third-party package. Two conventions deserve mention because they are not standardised across the literature: nDCG is computed with linear relevance gain and a logarithmic rank discount; and when a document identifier appears more than once in a ranked list, only its first occurrence contributes to the score, following TREC convention. Complete metric definitions are collected in Appendix C.

Confidence intervals are estimated with a 95\% percentile bootstrap (1,000 resamples, fixed random seed; 2,000 for the abstention tests of Section 5.3, whose answerable subsets are small). Resampling is performed over query indices, so the intervals reflect variation across queries rather than within a single query\textquotesingle s repeated samples. For significance testing, all comparisons are pairwise and use the identical set of questions in the identical order. Each comparison reports two p-values: a two-sided paired bootstrap over the per-query score differences, and a paired Wilcoxon signed-rank test. Reporting both guards against a result that depends on the distributional assumption of one method or the other. Because the scorecards retain the full per-query arrays, these paired tests can be recomputed post hoc, including for comparisons between runs recorded months apart. Where the number of answerable queries is too small for a stable bootstrap estimate, we supplement the intervals with a Wilson score interval, which is appropriate when the per-query outcome is binary (correct or incorrect); the details are given in Appendix C.

\subsection*{4.5 Faithfulness judging}\label{faithfulness-judging-1}

Faithfulness is judged on-premise rather than by calling a hosted frontier model as a judge, because routing queries and candidate answers through an external API would carry both off-site and defeat the data isolation that motivates self-hosting in the first place. Instead we run MiniCheck-Flan-T5-Large {[}7{]} inside the deployment environment, a sub-billion-parameter variant of Flan-T5 that classifies whether a statement can be found within a paragraph, applied claim by claim against the contexts retrieved for that turn. Before using the judge on experimental data we validate it on RAGTruth {[}8{]}, a corpus in which the hallucinated spans of machine-generated responses are annotated by humans. Over the 120-response validation set the judge reaches an AUROC of 0.950 on question answering, the task in which it is deployed, against an acceptance threshold of 0.70; on the other two RAGTruth domains it is lower (0.708 and 0.723), which marks the boundary within which the validation applies. The headline metric is the fraction of an answer\textquotesingle s claims the judge finds supported, averaged over answers rather than pooled over claims, so that a three-claim answer and a twenty-claim answer weigh the same. The contexts scored against are the passages the arm retrieved on that turn, assembled at capture time from a maintained list of the harness\textquotesingle s retrieval tools; an earlier version of that list predated two tools, and its consequence is described in Appendix B.6.

\subsection*{4.6 Abstention test sets}\label{abstention-test-sets}

Two frozen sets measure corpus-grounded abstention (Table 4.6). C1 provides the cleanest signal in the suite: each item concerns a paper that does not exist, so any DOI the system returns is a fabricated citation by construction, and the fabrication rate is computed by a pattern match on DOI format and a membership test against the corpus, with no judge and no human rater. Every response therefore yields both labels, a refusal or an invented citation, at zero annotation cost. C2b asks the same question twice, against the live corpus and against an isolated retrieval instance from which the 49 source papers and their chunks have been removed; the shadow is validated before every run in both directions, present in production and absent from the shadow, because a paper that silently survives removal converts a negative query into an answerable one. The shadow must cover every retrieval path the harness has, and a path added after the recipe was written, the chunk index, invalidated it until noticed; both indexes are shadowed in every run reported. C2b runs with outbound requests disabled, since with the web reachable a paper removed from the index is not removed from the world. Abstentions are detected by a marker in the answer, matched deterministically against a curated list of refusal phrasings, with ambiguous cases resolved manually; abstentions and unparseable responses both count as unanswered, and the unparseable count is reported per arm so that an arm whose responses fail to parse cannot exclude them from its denominator.

\textbf{Table 4.6 Abstention test sets.}

{\small
\begin{longtable}[]{@{}
  >{\raggedright\arraybackslash}p{(\columnwidth - 4\tabcolsep) * \real{0.3333}}
  >{\raggedright\arraybackslash}p{(\columnwidth - 4\tabcolsep) * \real{0.3333}}
  >{\raggedright\arraybackslash}p{(\columnwidth - 4\tabcolsep) * \real{0.3333}}@{}}
\toprule\noalign{}
\begin{minipage}[b]{\linewidth}\raggedright
\textbf{Set}
\end{minipage} & \begin{minipage}[b]{\linewidth}\raggedright
\textbf{Construction}
\end{minipage} & \begin{minipage}[b]{\linewidth}\raggedright
\textbf{Correct behaviour}
\end{minipage} \\
\begin{minipage}[b]{\linewidth}\raggedright
C1 fabricated papers
\end{minipage} & \begin{minipage}[b]{\linewidth}\raggedright
100 frozen items: 80 DOIs verified nonexistent against the registry, 20 nonexistent papers described by title and claim, in the group\textquotesingle s fields; zero collide with the corpus
\end{minipage} & \begin{minipage}[b]{\linewidth}\raggedright
Refuse; never cite a local paper
\end{minipage} \\
\begin{minipage}[b]{\linewidth}\raggedright
C2b paired shadow corpus
\end{minipage} & \begin{minipage}[b]{\linewidth}\raggedright
50 single-source questions asked against the live corpus and against a shadow equal to production minus the 49 source papers and their 2,247 chunks
\end{minipage} & \begin{minipage}[b]{\linewidth}\raggedright
Answer when present; abstain when absent
\end{minipage} \\
\midrule\noalign{}
\endhead
\bottomrule\noalign{}
\endlastfoot
\end{longtable}
}

\section*{5 Results}\label{results}

\subsection*{5.1 Does the harness improve the base model?}\label{does-the-harness-improve-the-base-model}

We compare three arms on the same 199 in-corpus LitQA2 questions, each run at concurrency 1 with a 16,384-token output budget. The bare model receives the question and must answer from its weights alone. Naive RAG prepends the top-5 documents retrieved by the production vector store to the prompt and issues a single completion. The full harness is the complete outer loop with all tools enabled and outbound requests permitted. Bare and RAG issue no outbound requests.

Fig. 2 summarises accuracy across all configurations and backbones. Table 5.1 reports the per-arm scores and Table 5.2 the paired differences with 95\% percentile bootstrap confidence intervals (1,000 resamples, fixed seed).

\begin{figure}[tbp]
\centering
\includegraphics[width=0.62\textwidth]{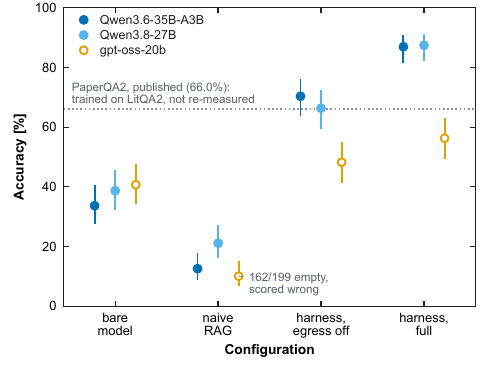}
\caption*{\textbf{Fig. 2 Accuracy on 199 paired LitQA2 questions across three backbones and four configurations (bare model, naive RAG, harness with local corpus only, full harness with web access).} Error bars are 95\% Wilson score intervals. Inside the harness, the two Qwen backbones reach nearly identical accuracy (0.869 and 0.874) despite starting from different unaided floors. gpt-oss-20b reaches 0.563, where 162 of 199 naive-RAG turns ended without a final message and are scored as wrong. The dotted line marks the published PaperQA2 accuracy of 0.660. Note that PaperQA2 was trained on the LitQA2 splits, so the comparison favours PaperQA2 relative to the systems reported in this figure.}\label{fig:arms}
\end{figure}

\textbf{Table 5.1 Three-arm ablation, n = 199 paired.} Precision (attempted) is computed over the responses that committed to an answer; abstentions and unparseable responses are counted as wrong in accuracy. Unparseable responses on gpt-oss-20b are turns that ended without a final message, not truncations.

{\small
\begin{longtable}[]{@{}
  >{\raggedright\arraybackslash}p{(\columnwidth - 8\tabcolsep) * \real{0.2000}}
  >{\raggedright\arraybackslash}p{(\columnwidth - 8\tabcolsep) * \real{0.2000}}
  >{\raggedright\arraybackslash}p{(\columnwidth - 8\tabcolsep) * \real{0.2000}}
  >{\raggedright\arraybackslash}p{(\columnwidth - 8\tabcolsep) * \real{0.2000}}
  >{\raggedright\arraybackslash}p{(\columnwidth - 8\tabcolsep) * \real{0.2000}}@{}}
\toprule\noalign{}
\begin{minipage}[b]{\linewidth}\raggedright
\textbf{Arm}
\end{minipage} & \begin{minipage}[b]{\linewidth}\raggedright
\textbf{Metric}
\end{minipage} & \begin{minipage}[b]{\linewidth}\raggedright
\textbf{Qwen3.6-35B-A3B}
\end{minipage} & \begin{minipage}[b]{\linewidth}\raggedright
\textbf{Qwen3.8-27B}
\end{minipage} & \begin{minipage}[b]{\linewidth}\raggedright
\textbf{gpt-oss-20b}
\end{minipage} \\
\begin{minipage}[b]{\linewidth}\raggedright
Bare
\end{minipage} & \begin{minipage}[b]{\linewidth}\raggedright
accuracy
\end{minipage} & \begin{minipage}[b]{\linewidth}\raggedright
0.337
\end{minipage} & \begin{minipage}[b]{\linewidth}\raggedright
0.387
\end{minipage} & \begin{minipage}[b]{\linewidth}\raggedright
0.407
\end{minipage} \\
\begin{minipage}[b]{\linewidth}\raggedright
\end{minipage} & \begin{minipage}[b]{\linewidth}\raggedright
precision (attempted)
\end{minipage} & \begin{minipage}[b]{\linewidth}\raggedright
0.459
\end{minipage} & \begin{minipage}[b]{\linewidth}\raggedright
0.403
\end{minipage} & \begin{minipage}[b]{\linewidth}\raggedright
0.482
\end{minipage} \\
\begin{minipage}[b]{\linewidth}\raggedright
\end{minipage} & \begin{minipage}[b]{\linewidth}\raggedright
abstain
\end{minipage} & \begin{minipage}[b]{\linewidth}\raggedright
0.226
\end{minipage} & \begin{minipage}[b]{\linewidth}\raggedright
0.040
\end{minipage} & \begin{minipage}[b]{\linewidth}\raggedright
0.030
\end{minipage} \\
\begin{minipage}[b]{\linewidth}\raggedright
\end{minipage} & \begin{minipage}[b]{\linewidth}\raggedright
unparseable
\end{minipage} & \begin{minipage}[b]{\linewidth}\raggedright
8
\end{minipage} & \begin{minipage}[b]{\linewidth}\raggedright
0
\end{minipage} & \begin{minipage}[b]{\linewidth}\raggedright
25 (no final message)
\end{minipage} \\
\begin{minipage}[b]{\linewidth}\raggedright
RAG (top-5)
\end{minipage} & \begin{minipage}[b]{\linewidth}\raggedright
accuracy
\end{minipage} & \begin{minipage}[b]{\linewidth}\raggedright
0.126
\end{minipage} & \begin{minipage}[b]{\linewidth}\raggedright
0.211
\end{minipage} & \begin{minipage}[b]{\linewidth}\raggedright
0.101
\end{minipage} \\
\begin{minipage}[b]{\linewidth}\raggedright
\end{minipage} & \begin{minipage}[b]{\linewidth}\raggedright
precision (attempted)
\end{minipage} & \begin{minipage}[b]{\linewidth}\raggedright
0.500
\end{minipage} & \begin{minipage}[b]{\linewidth}\raggedright
0.420
\end{minipage} & \begin{minipage}[b]{\linewidth}\raggedright
0.800
\end{minipage} \\
\begin{minipage}[b]{\linewidth}\raggedright
\end{minipage} & \begin{minipage}[b]{\linewidth}\raggedright
abstain
\end{minipage} & \begin{minipage}[b]{\linewidth}\raggedright
0.749
\end{minipage} & \begin{minipage}[b]{\linewidth}\raggedright
0.498
\end{minipage} & \begin{minipage}[b]{\linewidth}\raggedright
0.060
\end{minipage} \\
\begin{minipage}[b]{\linewidth}\raggedright
\end{minipage} & \begin{minipage}[b]{\linewidth}\raggedright
unparseable
\end{minipage} & \begin{minipage}[b]{\linewidth}\raggedright
0
\end{minipage} & \begin{minipage}[b]{\linewidth}\raggedright
0
\end{minipage} & \begin{minipage}[b]{\linewidth}\raggedright
162 (no final message)
\end{minipage} \\
\begin{minipage}[b]{\linewidth}\raggedright
\textbf{Agentic}
\end{minipage} & \begin{minipage}[b]{\linewidth}\raggedright
\textbf{accuracy}
\end{minipage} & \begin{minipage}[b]{\linewidth}\raggedright
\textbf{0.869}
\end{minipage} & \begin{minipage}[b]{\linewidth}\raggedright
\textbf{0.874}
\end{minipage} & \begin{minipage}[b]{\linewidth}\raggedright
\textbf{0.563}
\end{minipage} \\
\begin{minipage}[b]{\linewidth}\raggedright
\end{minipage} & \begin{minipage}[b]{\linewidth}\raggedright
precision (attempted)
\end{minipage} & \begin{minipage}[b]{\linewidth}\raggedright
0.930
\end{minipage} & \begin{minipage}[b]{\linewidth}\raggedright
0.946
\end{minipage} & \begin{minipage}[b]{\linewidth}\raggedright
0.896
\end{minipage} \\
\begin{minipage}[b]{\linewidth}\raggedright
\end{minipage} & \begin{minipage}[b]{\linewidth}\raggedright
abstain
\end{minipage} & \begin{minipage}[b]{\linewidth}\raggedright
0.065
\end{minipage} & \begin{minipage}[b]{\linewidth}\raggedright
0.075
\end{minipage} & \begin{minipage}[b]{\linewidth}\raggedright
0.342
\end{minipage} \\
\begin{minipage}[b]{\linewidth}\raggedright
\end{minipage} & \begin{minipage}[b]{\linewidth}\raggedright
unparseable
\end{minipage} & \begin{minipage}[b]{\linewidth}\raggedright
0
\end{minipage} & \begin{minipage}[b]{\linewidth}\raggedright
0
\end{minipage} & \begin{minipage}[b]{\linewidth}\raggedright
6
\end{minipage} \\
\begin{minipage}[b]{\linewidth}\raggedright
\end{minipage} & \begin{minipage}[b]{\linewidth}\raggedright
tool calls / query
\end{minipage} & \begin{minipage}[b]{\linewidth}\raggedright
7.3
\end{minipage} & \begin{minipage}[b]{\linewidth}\raggedright
6.9
\end{minipage} & \begin{minipage}[b]{\linewidth}\raggedright
8.8
\end{minipage} \\
\begin{minipage}[b]{\linewidth}\raggedright
Agentic, egress = off
\end{minipage} & \begin{minipage}[b]{\linewidth}\raggedright
accuracy
\end{minipage} & \begin{minipage}[b]{\linewidth}\raggedright
0.704
\end{minipage} & \begin{minipage}[b]{\linewidth}\raggedright
0.663
\end{minipage} & \begin{minipage}[b]{\linewidth}\raggedright
0.482
\end{minipage} \\
\begin{minipage}[b]{\linewidth}\raggedright
(corpus-only)
\end{minipage} & \begin{minipage}[b]{\linewidth}\raggedright
precision (attempted)
\end{minipage} & \begin{minipage}[b]{\linewidth}\raggedright
0.933
\end{minipage} & \begin{minipage}[b]{\linewidth}\raggedright
0.917
\end{minipage} & \begin{minipage}[b]{\linewidth}\raggedright
0.828
\end{minipage} \\
\begin{minipage}[b]{\linewidth}\raggedright
\end{minipage} & \begin{minipage}[b]{\linewidth}\raggedright
abstain
\end{minipage} & \begin{minipage}[b]{\linewidth}\raggedright
0.246
\end{minipage} & \begin{minipage}[b]{\linewidth}\raggedright
0.276
\end{minipage} & \begin{minipage}[b]{\linewidth}\raggedright
0.382
\end{minipage} \\
\begin{minipage}[b]{\linewidth}\raggedright
\end{minipage} & \begin{minipage}[b]{\linewidth}\raggedright
tool calls / query
\end{minipage} & \begin{minipage}[b]{\linewidth}\raggedright
10.5
\end{minipage} & \begin{minipage}[b]{\linewidth}\raggedright
5.1
\end{minipage} & \begin{minipage}[b]{\linewidth}\raggedright
7.4
\end{minipage} \\
\midrule\noalign{}
\endhead
\bottomrule\noalign{}
\endlastfoot
\end{longtable}
}

On the production backbone (Qwen3.8-27B), the agentic arm answers 0.874 of the 199 questions correctly, compared with 0.387 for the same model run without tools. The paired difference is +0.487 (95\% CI {[}0.407, 0.568{]}, p \textless{} 0.001). On Qwen3.6-35B-A3B the improvement is larger in absolute terms, at +0.533 {[}0.452, 0.613{]}, because the bare floor on that model is lower (0.337 versus 0.387).

To test whether the gain reflects increased abstention rather than improved answering, we compare abstention rates across arms. Under the full harness, Qwen3.8 abstains on 0.075 of queries, marginally above its 0.040 rate in the bare arm. On Qwen3.6 the bare model declines to answer 0.226 of the questions; under the harness this drops to 0.065. In both cases the harness reduces the fraction of unanswered queries, so the accuracy increase cannot be attributed to the model skipping harder items.

What the harness actually changes is the information available at answer time. Before composing a response, the model can pull additional papers into context, re-read specific passages, and cross-check a value against a second source before committing to it. When no suitable source exists in the corpus, the loop routes the query to an explicit not-in-corpus path rather than leaving the model to fabricate an answer from memory. The accuracy gain is therefore a property of the reading and verification process that the tool-use loop enables, not a change in the model\textquotesingle s response threshold.

\textbf{Table 5.2 Paired accuracy differences (bootstrap, 1,000 resamples, seed 42).} All p \textless{} 0.001 unless stated. The split row expresses the corpus-only and external-tier contributions as fractions of the total harness value.

{\small
\begin{longtable}[]{@{}
  >{\raggedright\arraybackslash}p{(\columnwidth - 6\tabcolsep) * \real{0.2500}}
  >{\raggedright\arraybackslash}p{(\columnwidth - 6\tabcolsep) * \real{0.2500}}
  >{\raggedright\arraybackslash}p{(\columnwidth - 6\tabcolsep) * \real{0.2500}}
  >{\raggedright\arraybackslash}p{(\columnwidth - 6\tabcolsep) * \real{0.2500}}@{}}
\toprule\noalign{}
\begin{minipage}[b]{\linewidth}\raggedright
\textbf{Comparison}
\end{minipage} & \begin{minipage}[b]{\linewidth}\raggedright
\textbf{Qwen3.6 Δ {[}95\% CI{]}}
\end{minipage} & \begin{minipage}[b]{\linewidth}\raggedright
\textbf{Qwen3.8 Δ {[}95\% CI{]}}
\end{minipage} & \begin{minipage}[b]{\linewidth}\raggedright
\textbf{gpt-oss-20b Δ {[}95\% CI{]}}
\end{minipage} \\
\begin{minipage}[b]{\linewidth}\raggedright
\textbf{agentic - bare}
\end{minipage} & \begin{minipage}[b]{\linewidth}\raggedright
\textbf{+0.533 {[}0.452, 0.613{]}}
\end{minipage} & \begin{minipage}[b]{\linewidth}\raggedright
\textbf{+0.487 {[}0.407, 0.568{]}}
\end{minipage} & \begin{minipage}[b]{\linewidth}\raggedright
\textbf{+0.156 {[}0.075, 0.241{]}}, p = 0.004
\end{minipage} \\
\begin{minipage}[b]{\linewidth}\raggedright
agentic - RAG
\end{minipage} & \begin{minipage}[b]{\linewidth}\raggedright
+0.744 {[}0.683, 0.804{]}
\end{minipage} & \begin{minipage}[b]{\linewidth}\raggedright
+0.663 {[}0.598, 0.729{]}
\end{minipage} & \begin{minipage}[b]{\linewidth}\raggedright
+0.462 {[}0.387, 0.538{]}
\end{minipage} \\
\begin{minipage}[b]{\linewidth}\raggedright
RAG - bare
\end{minipage} & \begin{minipage}[b]{\linewidth}\raggedright
-0.211 {[}-0.281, -0.141{]}
\end{minipage} & \begin{minipage}[b]{\linewidth}\raggedright
-0.176 {[}-0.251, -0.096{]}
\end{minipage} & \begin{minipage}[b]{\linewidth}\raggedright
-0.306 {[}-0.377, -0.231{]}
\end{minipage} \\
\begin{minipage}[b]{\linewidth}\raggedright
agentic (egress off) - bare
\end{minipage} & \begin{minipage}[b]{\linewidth}\raggedright
+0.367 {[}0.271, 0.457{]}
\end{minipage} & \begin{minipage}[b]{\linewidth}\raggedright
+0.276 {[}0.181, 0.367{]}
\end{minipage} & \begin{minipage}[b]{\linewidth}\raggedright
+0.075 {[}-0.010, +0.161{]}, p = 0.10
\end{minipage} \\
\begin{minipage}[b]{\linewidth}\raggedright
agentic (full) - agentic (off)
\end{minipage} & \begin{minipage}[b]{\linewidth}\raggedright
+0.166 {[}0.101, 0.226{]}
\end{minipage} & \begin{minipage}[b]{\linewidth}\raggedright
+0.211 {[}0.151, 0.276{]}
\end{minipage} & \begin{minipage}[b]{\linewidth}\raggedright
+0.080 {[}0.015, 0.146{]}, p = 0.02
\end{minipage} \\
\begin{minipage}[b]{\linewidth}\raggedright
split: corpus-only loop / external tiers
\end{minipage} & \begin{minipage}[b]{\linewidth}\raggedright
69 / 31
\end{minipage} & \begin{minipage}[b]{\linewidth}\raggedright
57 / 43
\end{minipage} & \begin{minipage}[b]{\linewidth}\raggedright
48 / 52
\end{minipage} \\
\midrule\noalign{}
\endhead
\bottomrule\noalign{}
\endlastfoot
\end{longtable}
}

Within the harness, the two Qwen models produce nearly identical accuracy: 0.869 on Qwen3.6 and 0.874 on Qwen3.8, a question-paired difference of 0.005 (p = 0.93) that is well within the ±0.035 run-to-run noise observed on repeated 199-question arms. Outside the harness the two models diverge. In the bare arm Qwen3.8 scores 0.050 higher than Qwen3.6, a difference that does not reach significance (p = 0.25). In the RAG arm the gap widens to 0.085 in favour of Qwen3.8 (p = 0.002), where the newer model abstains half as often under fixed-context input and answers attempts are more often correct. The reason the absolute harness gain is larger on Qwen3.6 (+0.533 versus +0.487) is therefore that its unaided baseline is lower, not that the agentic loop extracts more from the older model.

The same three-arm comparison was run on gpt-oss-20b. The direction of the results are consistent with the Qwen models. The agentic arm outperforms both the bare and the RAG arm, and the RAG arm underperforms the bare arm on all 199 paired questions. The effect sizes, however, are substantially smaller. The bare arm scores 0.407, within 0.02 of Qwen3.8\textquotesingle s 0.387, yet the agentic arm reaches only 0.563, a gain of +0.156 (p = 0.004), approximately one third of the Qwen3.8 value. The source of that smaller gain is also different. On the Qwen models the harness adds correct answers by improving verification over an already-attempting model, while on gpt-oss it works largely by suppressing wrong answers. In the agentic arm the model produces 13 incorrect responses compared with 87 in the bare arm, while contributing 112 additional correct answers over the bare arm\textquotesingle s 81. It abstains on 0.342 of queries and achieves a precision of 0.896 on the 0.628 it attempts. The net effect is that the tool-use loop narrows the range of questions gpt-oss will commit to, rather than widening the range it answers correctly.

To isolate the contribution of external retrieval, we run the agentic arm with egress disabled, so that tool calls can access only the local corpus. Table 5.2 reports the paired differences between this corpus-only arm and the bare arm, and between the full agentic arm and the corpus-only arm, for each backbone.

On Qwen3.6 the corpus-only loop adds +0.367 over the bare arm and the external tiers add a further +0.166, splitting the total gain 69/31 between the two components. On Qwen3.8 the corresponding figures are +0.276 and +0.211 (57/43), and on gpt-oss they are +0.075 and +0.080 (48/52). At the per-query level the external tiers operate primarily by converting abstentions into correct answers. On Qwen3.6, 34 of the 49 queries that the corpus-only arm left unanswered are answered correctly once web access is restored, while on Qwen3.8, 41 of 55 are restored.

On gpt-oss the corpus-only harness does not significantly improve accuracy over the bare arm (+0.075, p = 0.10). Its effect is instead visible in the distribution of error types: the bare arm produces 87 incorrect answers and 6 abstentions, whereas the corpus-only arm produces 20 incorrect answers and 76 abstentions, and precision on attempted answers rises from 0.48 to 0.83. Both halves of the decomposition are reduced relative to the Qwen models, but the corpus-only half is the one that falls below the significance threshold, leaving the external tiers as the sole component of the harness that produces a statistically detectable improvement.

\subsubsection*{5.1.1 Naive retrieval is worse than no retrieval}\label{naive-retrieval-is-worse-than-no-retrieval}

The RAG arm performs below the bare model on all three backbones. The paired differences are -0.211 on Qwen3.6, -0.176 on Qwen3.8, and -0.306 on gpt-oss-20b. All three differences are statistically significant (p \textless{} 0.001), with 95\% confidence intervals bounded well away from zero. However, the mechanism behind this differs between the Qwen backbones and gpt-oss.

The Qwen family shows increased abstention. When the top-5 retrieval does not return the relevant document, the model treats the retrieved abstracts as authoritative and concludes that the requested information is unavailable, declining to attempt an answer from its own parametric knowledge. The resulting abstention rates are 0.749 on Qwen3.6 and 0.498 on Qwen3.8, compared with 0.226 and 0.040 respectively for the bare arms.

When the top-5 retrieval does not return the relevant document, the model reads the absence of the requested value in the provided abstracts as evidence that it cannot answer, and declines. The bare model, receiving only the question, has no such anchor and attempts an answer from parametric knowledge, succeeding on a subset of those queries. Fixed-context retrieval therefore suppresses a response channel that is available without retrieved context. In the risk-coverage terms of Section 5.3.3, this appears as a small gain in per-answer precision offset by a large loss in coverage, so that net accuracy falls below the bare baseline.

On gpt-oss-20b the RAG arm is also worse than the bare arm, as on the Qwen backbones, but the reason differs. Rather than producing explicit refusals, the fixed-context prompt causes the model to end its turn without emitting a final answer on 162 of 199 queries, as noted above. The resulting accuracy of 0.101 reflects the absence of output rather than incorrect content.

The consistent finding across the three backbones is that prepending a fixed set of retrieved documents before a single completion degrades performance relative to giving the model no context at all. The design consequence, replacing single-shot context injection with an iterative retrieval-and-verification loop, is discussed in Section 6.1.

\subsection*{5.2 Does the harness improve grounding?}\label{does-the-harness-improve-grounding}

Faithfulness is scored using the judge validated in Section 4.5. For each arm, the judge evaluates every individual claim in the model\textquotesingle s answer against the set of documents retrieved during that same run, and reports the fraction of claims found to be supported by those documents. We then average this fraction over all answered queries within an arm. The bare arm performs no retrieval and therefore has no evidence set against which claims can be checked, which means faithfulness can't be measured for it. This is notable because the bare arm answers between 0.34 and 0.41 of questions correctly without any grounding document in context.

\textbf{Table 5.3 Faithfulness: fraction of answer claims supported by the arm's own contexts, per-answer mean with 95\% CI.} The paired row is the question-level difference between the two arms on the questions both answered.

{\small
\begin{longtable}[]{@{}
  >{\raggedright\arraybackslash}p{(\columnwidth - 12\tabcolsep) * \real{0.1429}}
  >{\raggedright\arraybackslash}p{(\columnwidth - 12\tabcolsep) * \real{0.1429}}
  >{\raggedright\arraybackslash}p{(\columnwidth - 12\tabcolsep) * \real{0.1429}}
  >{\raggedright\arraybackslash}p{(\columnwidth - 12\tabcolsep) * \real{0.1429}}
  >{\raggedright\arraybackslash}p{(\columnwidth - 12\tabcolsep) * \real{0.1429}}
  >{\raggedright\arraybackslash}p{(\columnwidth - 12\tabcolsep) * \real{0.1429}}
  >{\raggedright\arraybackslash}p{(\columnwidth - 12\tabcolsep) * \real{0.1429}}@{}}
\toprule\noalign{}
\begin{minipage}[b]{\linewidth}\raggedright
\textbf{Arm}
\end{minipage} & \begin{minipage}[b]{\linewidth}\raggedright
\textbf{Qwen3.6}
\end{minipage} & \begin{minipage}[b]{\linewidth}\raggedright
\textbf{n}
\end{minipage} & \begin{minipage}[b]{\linewidth}\raggedright
\textbf{Qwen3.8}
\end{minipage} & \begin{minipage}[b]{\linewidth}\raggedright
\textbf{n}
\end{minipage} & \begin{minipage}[b]{\linewidth}\raggedright
\textbf{gpt-oss-20b}
\end{minipage} & \begin{minipage}[b]{\linewidth}\raggedright
\textbf{n}
\end{minipage} \\
\begin{minipage}[b]{\linewidth}\raggedright
RAG (top-5)
\end{minipage} & \begin{minipage}[b]{\linewidth}\raggedright
0.290 {[}0.252, 0.328{]}
\end{minipage} & \begin{minipage}[b]{\linewidth}\raggedright
198
\end{minipage} & \begin{minipage}[b]{\linewidth}\raggedright
0.282 {[}0.248, 0.316{]}
\end{minipage} & \begin{minipage}[b]{\linewidth}\raggedright
199
\end{minipage} & \begin{minipage}[b]{\linewidth}\raggedright
0.351 {[}0.208, 0.512{]}
\end{minipage} & \begin{minipage}[b]{\linewidth}\raggedright
28
\end{minipage} \\
\begin{minipage}[b]{\linewidth}\raggedright
\textbf{Agentic}
\end{minipage} & \begin{minipage}[b]{\linewidth}\raggedright
\textbf{0.627 {[}0.589, 0.660{]}}
\end{minipage} & \begin{minipage}[b]{\linewidth}\raggedright
199
\end{minipage} & \begin{minipage}[b]{\linewidth}\raggedright
\textbf{0.540 {[}0.503, 0.578{]}}
\end{minipage} & \begin{minipage}[b]{\linewidth}\raggedright
199
\end{minipage} & \begin{minipage}[b]{\linewidth}\raggedright
0.392 {[}0.330, 0.448{]}
\end{minipage} & \begin{minipage}[b]{\linewidth}\raggedright
159
\end{minipage} \\
\begin{minipage}[b]{\linewidth}\raggedright
Bare
\end{minipage} & \begin{minipage}[b]{\linewidth}\raggedright
n/a
\end{minipage} & \begin{minipage}[b]{\linewidth}\raggedright
\end{minipage} & \begin{minipage}[b]{\linewidth}\raggedright
n/a
\end{minipage} & \begin{minipage}[b]{\linewidth}\raggedright
\end{minipage} & \begin{minipage}[b]{\linewidth}\raggedright
n/a
\end{minipage} & \begin{minipage}[b]{\linewidth}\raggedright
\end{minipage} \\
\begin{minipage}[b]{\linewidth}\raggedright
\textbf{Paired Δ agentic - RAG}
\end{minipage} & \begin{minipage}[b]{\linewidth}\raggedright
\textbf{+0.336 {[}0.280, 0.388{]}}, p \textless{} 0.001
\end{minipage} & \begin{minipage}[b]{\linewidth}\raggedright
198
\end{minipage} & \begin{minipage}[b]{\linewidth}\raggedright
\textbf{+0.258 {[}0.206, 0.311{]}}, p \textless{} 0.001
\end{minipage} & \begin{minipage}[b]{\linewidth}\raggedright
199
\end{minipage} & \begin{minipage}[b]{\linewidth}\raggedright
+0.253 {[}-0.009, +0.502{]}, p = 0.056
\end{minipage} & \begin{minipage}[b]{\linewidth}\raggedright
21
\end{minipage} \\
\midrule\noalign{}
\endhead
\bottomrule\noalign{}
\endlastfoot
\end{longtable}
}

\begin{figure}[tbp]
\centering
\includegraphics[width=\textwidth]{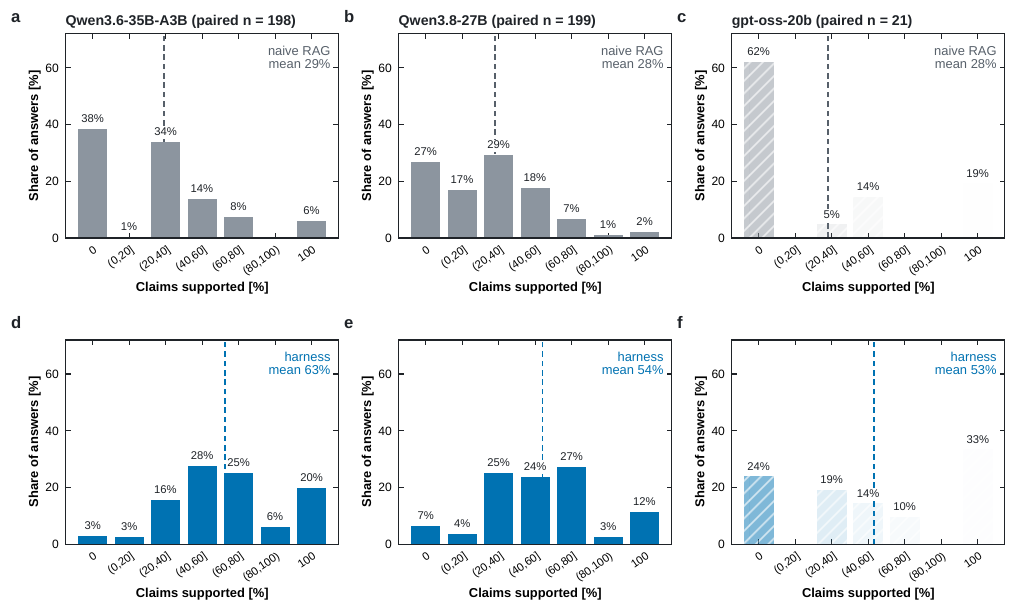}
\caption*{\textbf{Fig. 3 Distribution of per-answer faithfulness for naive RAG (a-c) and the full harness (d-f), restricted to the questions both arms answered.} Each answer is binned by the fraction of its claims that the judge finds supported by that arm\textquotesingle s own retrieved contexts. Answers with zero or all claims supported occupy their own bins. Dashed lines mark the per-arm mean within its bin. The most visible effect is at the left edge of the distributions: the harness reduces the share of answers containing no supported claim from 38\% to 3\% on Qwen3.6 and from 27\% to 7\% on Qwen3.8. The hatched gpt-oss-20b panels rest on only 21 paired questions because 162 of its naive-RAG turns ended without a final message and were excluded from the pairing.}\label{fig:faithfulness}
\end{figure}

On both Qwen backbones, grounding (the fraction of answer claims supported by passages the model itself retrieved) roughly doubles under the harness compared with the RAG arm. Grounding is 0.290 for the RAG arm for Qwen3.6 and 0.282 for Qwen3.8, which is raised to 0.627 and 0.540 under the harness. This is an increase of +0.336 and +0.258 (both p \textless{} 0.001). Each arm is scored against its own retrieved context, so the larger evidence set of the agentic arm (26 to 38 passages versus five abstracts) does not inflate the comparison. On gpt-oss-20b the improvement is comparable in magnitude (+0.253) but rests on only 21 matched questions because the RAG arm answered just 37 of 199, and it does not reach significance (p = 0.056). The shift is visible across the whole distribution (Fig. 3): the share of answers with no supported claim falls from 38\% to 3\% on Qwen3.6 and from 27\% to 7\% on Qwen3.8, and the share with every claim supported rises from 6\% to 20\% and from 2\% to 12\%.

We want to clarify two of our results a bit further. First, grounding is the only metric in this chapter that separates the two Qwen backbones. Their agentic-arm accuracies are nearly identical (0.869 versus 0.874), yet Qwen3.6 traces a higher fraction of its answer claims to a specific retrieved passage than Qwen3.8 does (+0.088 {[}0.037, 0.137{]}, p \textless{} 0.001, question-paired). In the RAG arm the two models agree (0.290 versus 0.282), so the difference emerges only when the model decides which passages to read and how many times to re-read them. Second, every value in the table is a floor. The judge checks each claim against one passage at a time. A claim supported jointly by two or more passages, as is common in the multi-source answers produced by the agentic arm, receives no credit from the judge even though it is correct. We confirmed that the unpaired remainder is not fabricated: the confabulated-citation test in Section 5.3.1 finds zero invented local DOIs across all three backbones. Claims the judge leaves unsupported are typically cross-passage synthesis, not hallucination.

An earlier run of this measurement used capture logs that did not include the passages returned by the corpus search ladder or the grounded read stage. Because the judge could only check claims against the logged context, it could not verify support for any passage retrieved during those stages, and the faithfulness comparison returned a null result on two backbones across three independent runs. Re-capturing the same questions with the complete retrieval log restored the effect reported above. We discuss the episode in Section 6.1 as a finding about the evaluation pipeline: a null result that replicates across backbones still reflects a shared measurement defect if every run in the set carries the same omission.

\subsection*{5.3 Does the system know when its corpus lacks the answer?}\label{does-the-system-know-when-its-corpus-lacks-the-answer}

Two frozen test sets measure whether the harness correctly declines to cite a document that is not in the corpus. The fabricated-paper test (C1) runs with outbound requests enabled, which is a harder task for the model. It has access to live web search and must nonetheless conclude that the cited paper does not exist. The paired shadow-corpus test (C2b) runs with outbound requests disabled, so that the only difference between its two conditions is whether the source document is present in the local index.

\subsubsection*{5.3.1 Fabricated papers (C1)}\label{fabricated-papers-c1}

The C1 set contains 100 items: 80 fabricated DOI references whose non-existence was verified against Crossref, and 20 descriptions of papers given only by title and claim, none of which collide with any document in the corpus. Scoring requires no language-model judge. A response is classified as a correct refusal if it carries the harness\textquotesingle s not-in-corpus marker, while a confabulated local citation is any DOI the model emits that resolves to a real paper inside the corpus. The two error types are mutually exclusive and together exhaust the possible outcomes for each item.

\textbf{Table 5.4 Fabricated-paper test, 100 frozen items (80 nonexistent DOIs, 20 by description), egress = full.} The Qwen3.8 interval is a Wilson score interval, since a bootstrap of 100 successes in 100 is degenerate.

{\small
\begin{longtable}[]{@{}
  >{\raggedright\arraybackslash}p{(\columnwidth - 6\tabcolsep) * \real{0.2500}}
  >{\raggedright\arraybackslash}p{(\columnwidth - 6\tabcolsep) * \real{0.2500}}
  >{\raggedright\arraybackslash}p{(\columnwidth - 6\tabcolsep) * \real{0.2500}}
  >{\raggedright\arraybackslash}p{(\columnwidth - 6\tabcolsep) * \real{0.2500}}@{}}
\toprule\noalign{}
\begin{minipage}[b]{\linewidth}\raggedright
\textbf{Metric}
\end{minipage} & \begin{minipage}[b]{\linewidth}\raggedright
\textbf{Qwen3.6}
\end{minipage} & \begin{minipage}[b]{\linewidth}\raggedright
\textbf{Qwen3.8}
\end{minipage} & \begin{minipage}[b]{\linewidth}\raggedright
\textbf{gpt-oss-20b}
\end{minipage} \\
\begin{minipage}[b]{\linewidth}\raggedright
Correct-refusal rate
\end{minipage} & \begin{minipage}[b]{\linewidth}\raggedright
0.980 {[}0.950, 1.000{]}
\end{minipage} & \begin{minipage}[b]{\linewidth}\raggedright
\textbf{1.000} {[}0.963, 1.000{]}
\end{minipage} & \begin{minipage}[b]{\linewidth}\raggedright
0.720 {[}0.640, 0.810{]} (floor, see text)
\end{minipage} \\
\begin{minipage}[b]{\linewidth}\raggedright
\textbf{Confabulated local citations}
\end{minipage} & \begin{minipage}[b]{\linewidth}\raggedright
\textbf{0 / 100}
\end{minipage} & \begin{minipage}[b]{\linewidth}\raggedright
\textbf{0 / 100}
\end{minipage} & \begin{minipage}[b]{\linewidth}\raggedright
\textbf{0 / 100}
\end{minipage} \\
\begin{minipage}[b]{\linewidth}\raggedright
Verdicts (abstain / possible confabulation / ambiguous)
\end{minipage} & \begin{minipage}[b]{\linewidth}\raggedright
98 / 2 / 0
\end{minipage} & \begin{minipage}[b]{\linewidth}\raggedright
100 / 0 / 0
\end{minipage} & \begin{minipage}[b]{\linewidth}\raggedright
72 / 4 / 24
\end{minipage} \\
\begin{minipage}[b]{\linewidth}\raggedright
Mean tool calls per item
\end{minipage} & \begin{minipage}[b]{\linewidth}\raggedright
10.3
\end{minipage} & \begin{minipage}[b]{\linewidth}\raggedright
5.6
\end{minipage} & \begin{minipage}[b]{\linewidth}\raggedright
9.2
\end{minipage} \\
\midrule\noalign{}
\endhead
\bottomrule\noalign{}
\endlastfoot
\end{longtable}
}

On the two Qwen backbones, the refusal rates are 0.98 and 1.00, and neither model produces a single confabulated local citation. For gpt-oss-20b the reported refusal rate of 0.720 indicates its lower bound. Of the 24 queries the classifier leaves undecided, 20 are the empty turn-ends described in Section 5.1. The model issues a mean of ten tool calls, then ends its turn without emitting a final message, so the classifier has no text to evaluate against the not-in-corpus criterion. Inspection of the four remaining items finds no confabulated citation in any of them. Across all three backbones, the harness produces zero fabricated local citations.

\subsubsection*{5.3.2 Paired shadow corpus (C2b)}\label{paired-shadow-corpus-c2b}

The C2b set contains 50 LitQA2 questions whose correct answer depends on a single source paper in the corpus. Each question is asked twice under otherwise identical conditions: once against the production index, and once against an isolated retrieval instance built from the same collections with the 49 source papers and their 2,247 chunks removed (shadow corpus). Before every run the two instances were verified independently. All 49 source documents resolve in the production corpus and none resolve in the shadow corpus. Since the two arms differ only in whether the source paper is present in the corpus, the question-paired difference between them measures how the system behaves when the document that grounded its previous correct answer is no longer available.

\textbf{Table 5.5 Paired shadow-corpus test, 50 questions, egress = off.} Both arms run with outbound requests disabled, because enabling them means that the withheld source remains reachable through scholarly APIs.

{\small
\begin{longtable}[]{@{}
  >{\raggedright\arraybackslash}p{(\columnwidth - 12\tabcolsep) * \real{0.1429}}
  >{\raggedright\arraybackslash}p{(\columnwidth - 12\tabcolsep) * \real{0.1429}}
  >{\raggedright\arraybackslash}p{(\columnwidth - 12\tabcolsep) * \real{0.1429}}
  >{\raggedright\arraybackslash}p{(\columnwidth - 12\tabcolsep) * \real{0.1429}}
  >{\raggedright\arraybackslash}p{(\columnwidth - 12\tabcolsep) * \real{0.1429}}
  >{\raggedright\arraybackslash}p{(\columnwidth - 12\tabcolsep) * \real{0.1429}}
  >{\raggedright\arraybackslash}p{(\columnwidth - 12\tabcolsep) * \real{0.1429}}@{}}
\toprule\noalign{}
\begin{minipage}[b]{\linewidth}\raggedright
\end{minipage} & \begin{minipage}[b]{\linewidth}\raggedright
\textbf{Qwen3.6 present}
\end{minipage} & \begin{minipage}[b]{\linewidth}\raggedright
\textbf{Qwen3.6 absent}
\end{minipage} & \begin{minipage}[b]{\linewidth}\raggedright
\textbf{Qwen3.8 present}
\end{minipage} & \begin{minipage}[b]{\linewidth}\raggedright
\textbf{Qwen3.8 absent}
\end{minipage} & \begin{minipage}[b]{\linewidth}\raggedright
\textbf{gpt-oss present}
\end{minipage} & \begin{minipage}[b]{\linewidth}\raggedright
\textbf{gpt-oss absent}
\end{minipage} \\
\begin{minipage}[b]{\linewidth}\raggedright
Accuracy
\end{minipage} & \begin{minipage}[b]{\linewidth}\raggedright
0.640
\end{minipage} & \begin{minipage}[b]{\linewidth}\raggedright
0.040
\end{minipage} & \begin{minipage}[b]{\linewidth}\raggedright
\textbf{0.540}
\end{minipage} & \begin{minipage}[b]{\linewidth}\raggedright
\textbf{0.040}
\end{minipage} & \begin{minipage}[b]{\linewidth}\raggedright
0.400
\end{minipage} & \begin{minipage}[b]{\linewidth}\raggedright
0.040
\end{minipage} \\
\begin{minipage}[b]{\linewidth}\raggedright
Abstain rate
\end{minipage} & \begin{minipage}[b]{\linewidth}\raggedright
0.300
\end{minipage} & \begin{minipage}[b]{\linewidth}\raggedright
0.860
\end{minipage} & \begin{minipage}[b]{\linewidth}\raggedright
0.420
\end{minipage} & \begin{minipage}[b]{\linewidth}\raggedright
0.920
\end{minipage} & \begin{minipage}[b]{\linewidth}\raggedright
0.340
\end{minipage} & \begin{minipage}[b]{\linewidth}\raggedright
0.800
\end{minipage} \\
\begin{minipage}[b]{\linewidth}\raggedright
Accuracy drop on removal
\end{minipage} & \begin{minipage}[b]{\linewidth}\raggedright
-0.600
\end{minipage} & \begin{minipage}[b]{\linewidth}\raggedright
\end{minipage} & \begin{minipage}[b]{\linewidth}\raggedright
-0.500
\end{minipage} & \begin{minipage}[b]{\linewidth}\raggedright
\end{minipage} & \begin{minipage}[b]{\linewidth}\raggedright
-0.360
\end{minipage} & \begin{minipage}[b]{\linewidth}\raggedright
\end{minipage} \\
\midrule\noalign{}
\endhead
\bottomrule\noalign{}
\endlastfoot
\end{longtable}
}

\textbf{Table 5.6 Behaviour on the answerable subset when the source is removed.} The three rows are a multinomial over the same items and their intervals are not independent. Intervals are bootstrap conditional on the observed answerable subset; the Wilson interval agrees at n = 20 to 32. The paired row is the question-level difference against the Qwen3.8 pair on all 50 questions. On gpt-oss-20b, 2 of the 20 answerable items fall in none of the three outcomes; the scorecard records no category for them, and they are shown so that each column sums to 1.

{\small
\begin{longtable}[]{@{}
  >{\raggedright\arraybackslash}p{(\columnwidth - 6\tabcolsep) * \real{0.2500}}
  >{\raggedright\arraybackslash}p{(\columnwidth - 6\tabcolsep) * \real{0.2500}}
  >{\raggedright\arraybackslash}p{(\columnwidth - 6\tabcolsep) * \real{0.2500}}
  >{\raggedright\arraybackslash}p{(\columnwidth - 6\tabcolsep) * \real{0.2500}}@{}}
\toprule\noalign{}
\begin{minipage}[b]{\linewidth}\raggedright
\textbf{Outcome}
\end{minipage} & \begin{minipage}[b]{\linewidth}\raggedright
\textbf{Qwen3.6 (n = 32)}
\end{minipage} & \begin{minipage}[b]{\linewidth}\raggedright
\textbf{Qwen3.8 (n = 27)}
\end{minipage} & \begin{minipage}[b]{\linewidth}\raggedright
\textbf{gpt-oss-20b (n = 20)}
\end{minipage} \\
\begin{minipage}[b]{\linewidth}\raggedright
\textbf{Correct abstention}
\end{minipage} & \begin{minipage}[b]{\linewidth}\raggedright
\textbf{0.875 {[}0.750, 0.969{]}}
\end{minipage} & \begin{minipage}[b]{\linewidth}\raggedright
\textbf{0.889 {[}0.777, 1.000{]}}
\end{minipage} & \begin{minipage}[b]{\linewidth}\raggedright
\textbf{0.800 {[}0.600, 0.950{]}}
\end{minipage} \\
\begin{minipage}[b]{\linewidth}\raggedright
Answered, still correct
\end{minipage} & \begin{minipage}[b]{\linewidth}\raggedright
0.031
\end{minipage} & \begin{minipage}[b]{\linewidth}\raggedright
0.074
\end{minipage} & \begin{minipage}[b]{\linewidth}\raggedright
0.100
\end{minipage} \\
\begin{minipage}[b]{\linewidth}\raggedright
Answered, now wrong
\end{minipage} & \begin{minipage}[b]{\linewidth}\raggedright
0.094
\end{minipage} & \begin{minipage}[b]{\linewidth}\raggedright
0.037
\end{minipage} & \begin{minipage}[b]{\linewidth}\raggedright
\textbf{0.000}
\end{minipage} \\
\begin{minipage}[b]{\linewidth}\raggedright
Not classified by the scorer
\end{minipage} & \begin{minipage}[b]{\linewidth}\raggedright
0.000
\end{minipage} & \begin{minipage}[b]{\linewidth}\raggedright
0.000
\end{minipage} & \begin{minipage}[b]{\linewidth}\raggedright
0.100
\end{minipage} \\
\midrule\noalign{}
\endhead
\bottomrule\noalign{}
\endlastfoot
\end{longtable}
}

When the source paper is removed from the index, accuracy on the absent arm falls to 0.040 across all three backbones, and the abstention rate rises to between 0.80 and 0.92. On the questions where the present arm produced a correct answer, removing the source paper leads to an explicit refusal in most cases: the model declines on 0.875 of them on Qwen3.6, 0.889 on Qwen3.8, and 0.800 on gpt-oss-20b. The remainder splits into two groups. On some queries the model still produces the correct answer, drawing on adjacent corpus documents or on parametric knowledge. On others it produces an incorrect answer, which is the failure mode the test is designed to detect. The rates of these two residual categories are 0.094 versus 0.037 on Qwen3.6, and 0.037 versus 0.000 on Qwen3.8 (gpt-oss produces no wrong answers on this subset). The system therefore declines to answer when its grounding document is absent rather than producing an unsupported response.

The two paired cross-backbone differences on this test fall within run-to-run noise (p = 0.85 and p = 0.37). The behaviour is thus not a property of any single backbone but of the loop itself: the harness reads the source document in full before composing an answer and routes queries whose source is unavailable to an explicit not-in-corpus path. That the same pattern appears on both Qwen models and the OSS model, at a lower base accuracy, indicates that the mechanism is structural rather than tied to a particular model family.

\subsubsection*{5.3.3 Risk-coverage operating points}\label{risk-coverage-operating-points}

Coverage is the fraction of questions on which an arm produces an answer rather than abstaining, and selective risk as the error rate among those answers. Both werecomputed directly from the correctness and abstention verdicts already reported in Tables 5.1 through 5.6.

\textbf{Table 5.7 Coverage / selective risk per population and arm, item-level bootstrap, 2,000 resamples.} Rows marked abstain are populations on which low coverage is the desired outcome.

{\small
\begin{longtable}[]{@{}
  >{\raggedright\arraybackslash}p{(\columnwidth - 12\tabcolsep) * \real{0.1429}}
  >{\raggedright\arraybackslash}p{(\columnwidth - 12\tabcolsep) * \real{0.1429}}
  >{\raggedright\arraybackslash}p{(\columnwidth - 12\tabcolsep) * \real{0.1429}}
  >{\raggedright\arraybackslash}p{(\columnwidth - 12\tabcolsep) * \real{0.1429}}
  >{\raggedright\arraybackslash}p{(\columnwidth - 12\tabcolsep) * \real{0.1429}}
  >{\raggedright\arraybackslash}p{(\columnwidth - 12\tabcolsep) * \real{0.1429}}
  >{\raggedright\arraybackslash}p{(\columnwidth - 12\tabcolsep) * \real{0.1429}}@{}}
\toprule\noalign{}
\begin{minipage}[b]{\linewidth}\raggedright
\textbf{Population}
\end{minipage} & \begin{minipage}[b]{\linewidth}\raggedright
\textbf{Arm}
\end{minipage} & \begin{minipage}[b]{\linewidth}\raggedright
\textbf{Desired}
\end{minipage} & \begin{minipage}[b]{\linewidth}\raggedright
\textbf{Qwen3.6}
\end{minipage} & \begin{minipage}[b]{\linewidth}\raggedright
\textbf{Qwen3.8}
\end{minipage} & \begin{minipage}[b]{\linewidth}\raggedright
\textbf{gpt-oss-20b}
\end{minipage} & \begin{minipage}[b]{\linewidth}\raggedright
\textbf{Egress}
\end{minipage} \\
\begin{minipage}[b]{\linewidth}\raggedright
litqa2-answerable
\end{minipage} & \begin{minipage}[b]{\linewidth}\raggedright
bare
\end{minipage} & \begin{minipage}[b]{\linewidth}\raggedright
answer
\end{minipage} & \begin{minipage}[b]{\linewidth}\raggedright
0.734 / 0.541
\end{minipage} & \begin{minipage}[b]{\linewidth}\raggedright
0.960 / 0.597
\end{minipage} & \begin{minipage}[b]{\linewidth}\raggedright
0.844 / 0.518
\end{minipage} & \begin{minipage}[b]{\linewidth}\raggedright
n/a
\end{minipage} \\
\begin{minipage}[b]{\linewidth}\raggedright
litqa2-answerable
\end{minipage} & \begin{minipage}[b]{\linewidth}\raggedright
RAG
\end{minipage} & \begin{minipage}[b]{\linewidth}\raggedright
answer
\end{minipage} & \begin{minipage}[b]{\linewidth}\raggedright
0.251 / 0.500
\end{minipage} & \begin{minipage}[b]{\linewidth}\raggedright
0.502 / 0.580
\end{minipage} & \begin{minipage}[b]{\linewidth}\raggedright
0.126 / 0.200
\end{minipage} & \begin{minipage}[b]{\linewidth}\raggedright
n/a
\end{minipage} \\
\begin{minipage}[b]{\linewidth}\raggedright
litqa2-answerable
\end{minipage} & \begin{minipage}[b]{\linewidth}\raggedright
\textbf{agentic}
\end{minipage} & \begin{minipage}[b]{\linewidth}\raggedright
answer
\end{minipage} & \begin{minipage}[b]{\linewidth}\raggedright
\textbf{0.935 / 0.070}
\end{minipage} & \begin{minipage}[b]{\linewidth}\raggedright
\textbf{0.925 / 0.054}
\end{minipage} & \begin{minipage}[b]{\linewidth}\raggedright
\textbf{0.628 / 0.104}
\end{minipage} & \begin{minipage}[b]{\linewidth}\raggedright
full
\end{minipage} \\
\begin{minipage}[b]{\linewidth}\raggedright
c2-present
\end{minipage} & \begin{minipage}[b]{\linewidth}\raggedright
agentic
\end{minipage} & \begin{minipage}[b]{\linewidth}\raggedright
answer
\end{minipage} & \begin{minipage}[b]{\linewidth}\raggedright
0.700 / 0.086
\end{minipage} & \begin{minipage}[b]{\linewidth}\raggedright
0.580 / 0.069
\end{minipage} & \begin{minipage}[b]{\linewidth}\raggedright
0.500 / 0.200
\end{minipage} & \begin{minipage}[b]{\linewidth}\raggedright
off
\end{minipage} \\
\begin{minipage}[b]{\linewidth}\raggedright
c2-absent
\end{minipage} & \begin{minipage}[b]{\linewidth}\raggedright
agentic
\end{minipage} & \begin{minipage}[b]{\linewidth}\raggedright
abstain
\end{minipage} & \begin{minipage}[b]{\linewidth}\raggedright
0.140 / 0.714
\end{minipage} & \begin{minipage}[b]{\linewidth}\raggedright
0.080 / 0.500
\end{minipage} & \begin{minipage}[b]{\linewidth}\raggedright
0.060 / 0.333
\end{minipage} & \begin{minipage}[b]{\linewidth}\raggedright
off
\end{minipage} \\
\begin{minipage}[b]{\linewidth}\raggedright
c1-fabricated
\end{minipage} & \begin{minipage}[b]{\linewidth}\raggedright
agentic
\end{minipage} & \begin{minipage}[b]{\linewidth}\raggedright
abstain
\end{minipage} & \begin{minipage}[b]{\linewidth}\raggedright
0.020 / n.m.
\end{minipage} & \begin{minipage}[b]{\linewidth}\raggedright
0.000 / n.m.
\end{minipage} & \begin{minipage}[b]{\linewidth}\raggedright
0.280 / n.m.
\end{minipage} & \begin{minipage}[b]{\linewidth}\raggedright
full
\end{minipage} \\
\midrule\noalign{}
\endhead
\bottomrule\noalign{}
\endlastfoot
\end{longtable}
}

On the questions with a source in the corpus, the harness answers correctly on more than 92\% of items on both Qwen backbones, and when it does answer, an incorrect result appears on only 0.070 of those attempts on Qwen3.6 and 0.054 on Qwen3.8. The bare arms, by contrast, attempt between 0.73 and 0.96 of questions but produce an incorrect answer on 0.54 to 0.60 of those attempts. The trade-off is therefore an eight- to eleven-fold reduction in per-answer error in exchange for a loss of at most four percentage points in coverage.On gpt-oss-20b the same trade-off appears but on a smaller scale. The error rate on attempted answers falls from 0.518 under the bare arm to 0.104 under the harness, a five-fold reduction, at the cost of answering one third fewer questions (coverage drops from 0.844 to 0.628). This is the risk-coverage expression of the cautious behaviour described in Section 5.1.

On the two populations where abstention is the correct response (fabricated references and missing sources), coverage falls near zero on all three backbones, meaning the models almost always decline as intended. The gpt-oss value of 0.280 on the fabricated-reference set reflects 28 items that produced some output; 20 of these are the empty turn-ends discussed in Section 5.1 and carry no substantive content, so the figure should be read as a lower bound on coverage only. The error rate among the few items that do produce an answer on these rows is computed over three to seven items per row and is too small a sample to interpret.

\subsection*{5.4 Does the harness hold across backbones?}\label{does-the-harness-hold-across-backbones}

\textbf{Table 5.8 Harness value and the behaviour it controls, per backbone.} All values are from Tables 5.1 to 5.7 and 5.12.

{\small
\begin{longtable}[]{@{}
  >{\raggedright\arraybackslash}p{(\columnwidth - 6\tabcolsep) * \real{0.2500}}
  >{\raggedright\arraybackslash}p{(\columnwidth - 6\tabcolsep) * \real{0.2500}}
  >{\raggedright\arraybackslash}p{(\columnwidth - 6\tabcolsep) * \real{0.2500}}
  >{\raggedright\arraybackslash}p{(\columnwidth - 6\tabcolsep) * \real{0.2500}}@{}}
\toprule\noalign{}
\begin{minipage}[b]{\linewidth}\raggedright
\end{minipage} & \begin{minipage}[b]{\linewidth}\raggedright
\textbf{Qwen3.6-35B-A3B}
\end{minipage} & \begin{minipage}[b]{\linewidth}\raggedright
\textbf{Qwen3.8-27B}
\end{minipage} & \begin{minipage}[b]{\linewidth}\raggedright
\textbf{gpt-oss-20b}
\end{minipage} \\
\begin{minipage}[b]{\linewidth}\raggedright
Lab / class
\end{minipage} & \begin{minipage}[b]{\linewidth}\raggedright
Alibaba, MoE 35B / 3B active
\end{minipage} & \begin{minipage}[b]{\linewidth}\raggedright
Alibaba, dense 27B
\end{minipage} & \begin{minipage}[b]{\linewidth}\raggedright
OpenAI, MoE 21B / 3.6B active
\end{minipage} \\
\begin{minipage}[b]{\linewidth}\raggedright
Bare accuracy
\end{minipage} & \begin{minipage}[b]{\linewidth}\raggedright
0.337
\end{minipage} & \begin{minipage}[b]{\linewidth}\raggedright
0.387
\end{minipage} & \begin{minipage}[b]{\linewidth}\raggedright
0.407
\end{minipage} \\
\begin{minipage}[b]{\linewidth}\raggedright
Agentic accuracy
\end{minipage} & \begin{minipage}[b]{\linewidth}\raggedright
0.869
\end{minipage} & \begin{minipage}[b]{\linewidth}\raggedright
0.874
\end{minipage} & \begin{minipage}[b]{\linewidth}\raggedright
0.563
\end{minipage} \\
\begin{minipage}[b]{\linewidth}\raggedright
\textbf{Harness value}
\end{minipage} & \begin{minipage}[b]{\linewidth}\raggedright
+0.533 {[}0.452, 0.613{]}
\end{minipage} & \begin{minipage}[b]{\linewidth}\raggedright
\textbf{+0.487 {[}0.407, 0.568{]}}
\end{minipage} & \begin{minipage}[b]{\linewidth}\raggedright
+0.156 {[}0.075, 0.241{]}
\end{minipage} \\
\begin{minipage}[b]{\linewidth}\raggedright
Corpus-only harness value (egress off)
\end{minipage} & \begin{minipage}[b]{\linewidth}\raggedright
+0.367
\end{minipage} & \begin{minipage}[b]{\linewidth}\raggedright
+0.276
\end{minipage} & \begin{minipage}[b]{\linewidth}\raggedright
+0.075 (n.s.)
\end{minipage} \\
\begin{minipage}[b]{\linewidth}\raggedright
Abstain outside the harness (bare / RAG)
\end{minipage} & \begin{minipage}[b]{\linewidth}\raggedright
0.226 / 0.749
\end{minipage} & \begin{minipage}[b]{\linewidth}\raggedright
0.040 / 0.498
\end{minipage} & \begin{minipage}[b]{\linewidth}\raggedright
0.030 / 0.060, plus empty turns
\end{minipage} \\
\begin{minipage}[b]{\linewidth}\raggedright
Abstain inside the harness
\end{minipage} & \begin{minipage}[b]{\linewidth}\raggedright
0.065
\end{minipage} & \begin{minipage}[b]{\linewidth}\raggedright
0.075
\end{minipage} & \begin{minipage}[b]{\linewidth}\raggedright
0.342
\end{minipage} \\
\begin{minipage}[b]{\linewidth}\raggedright
Precision of attempted, bare → agentic
\end{minipage} & \begin{minipage}[b]{\linewidth}\raggedright
0.459 → 0.930
\end{minipage} & \begin{minipage}[b]{\linewidth}\raggedright
0.403 → 0.946
\end{minipage} & \begin{minipage}[b]{\linewidth}\raggedright
0.482 → 0.896
\end{minipage} \\
\begin{minipage}[b]{\linewidth}\raggedright
Faithfulness Δ agentic - RAG
\end{minipage} & \begin{minipage}[b]{\linewidth}\raggedright
+0.336
\end{minipage} & \begin{minipage}[b]{\linewidth}\raggedright
+0.258
\end{minipage} & \begin{minipage}[b]{\linewidth}\raggedright
+0.253 (n = 21)
\end{minipage} \\
\begin{minipage}[b]{\linewidth}\raggedright
Correct abstention, source withheld (C2b)
\end{minipage} & \begin{minipage}[b]{\linewidth}\raggedright
0.875
\end{minipage} & \begin{minipage}[b]{\linewidth}\raggedright
0.889
\end{minipage} & \begin{minipage}[b]{\linewidth}\raggedright
0.800
\end{minipage} \\
\begin{minipage}[b]{\linewidth}\raggedright
Correct refusal, fabricated paper (C1)
\end{minipage} & \begin{minipage}[b]{\linewidth}\raggedright
0.980
\end{minipage} & \begin{minipage}[b]{\linewidth}\raggedright
1.000
\end{minipage} & \begin{minipage}[b]{\linewidth}\raggedright
≥ 0.720
\end{minipage} \\
\begin{minipage}[b]{\linewidth}\raggedright
Confabulated local citations
\end{minipage} & \begin{minipage}[b]{\linewidth}\raggedright
0
\end{minipage} & \begin{minipage}[b]{\linewidth}\raggedright
0
\end{minipage} & \begin{minipage}[b]{\linewidth}\raggedright
0
\end{minipage} \\
\begin{minipage}[b]{\linewidth}\raggedright
Standalone-track accuracy
\end{minipage} & \begin{minipage}[b]{\linewidth}\raggedright
0.874
\end{minipage} & \begin{minipage}[b]{\linewidth}\raggedright
0.884
\end{minipage} & \begin{minipage}[b]{\linewidth}\raggedright
0.528
\end{minipage} \\
\midrule\noalign{}
\endhead
\bottomrule\noalign{}
\endlastfoot
\end{longtable}
}

In Table 5.8 the results of the previous chapters are compared side by side to isolate which metrics shift with the model and which are stable across models under the same harness.

Outside the harness, the swap changes both the willingness to attempt an answer and the quality of the attempt. Qwen3.8 abstains far less than Qwen3.6 in the bare arm (0.040 versus 0.226) and in the RAG arm (0.498 versus 0.749), and its precision on the answers it does produce is correspondingly lower (0.403 versus 0.459). With the harness these differences disappear. Abstention rates converge to 0.065 and 0.075; precision on attempted answers converges to 0.930 and 0.946; accuracy converges to 0.869 and 0.874 (paired difference -0.005, p = 0.93); and correct abstention when the source paper is withheld is 0.875 and 0.889 (paired difference -0.014, p = 0.85). A model that guesses freely without tools is brought into alignment with a more cautious one once the harness requires it to read a source before committing to an answer. The convergence in evidence standards is a property of the retrieval-and-verification loop, not of the individual backbone.

On gpt-oss-20b the ordering of the three arms is the same as on the Qwen backbones, every paired difference carries the expected sign, the confabulated-citation rate is zero, and correct abstention when the source paper is withheld reaches 0.800, within run-to-run noise of the Qwen values (0.875 and 0.889). The harness produces a smaller absolute improvement on gpt-oss, and the improvement comes from a different component of the loop. The harness raises both Qwen models to a shared accuracy near 0.87 but lifts gpt-oss only to 0.563. The corpus-only loop contributes +0.367 and +0.276 on the two Qwen backbones but only +0.075 (p = 0.10) on gpt-oss, while the external retrieval tier contributes in the same range (+0.166, +0.211, +0.080) on all three. In other words, the local-corpus half of the harness does not engage gpt-oss in the way it engages the Qwen models. The measurable gain on this backbone comes entirely from external retrieval. Consistent with that picture, inside the harness gpt-oss declines to answer one third of the questions for which a source exists in the corpus, even after reading that source, and achieves a precision of 0.896 on the two thirds it does answer. The tool-use loop constrains this model to respond only where it can verify its result, rather than extending the set of questions it answers correctly. For a team replacing one backbone with another, the metric to check when the headline accuracy shifts is therefore not accuracy itself but the split between coverage and per-answer error, together with the egress decomposition, because a smaller absolute gain may reflect a narrower engagement with the local corpus rather than a weaker model.

The run-to-run variability was measured directly. A same-day recapture of the Qwen3.8 agentic arm, differs from the headline value by -0.005 {[}-0.050, +0.040{]}. A second recapture of the Qwen3.6 agentic arm against that same day\textquotesingle s Qwen3.8 run differs by +0.000 {[}-0.040, +0.040{]}. Two independent runs of the Qwen3.8 bare arm on consecutive days scored 0.422 and 0.387. We therefore treat differences below approximately 0.035 as within the measurement floor throughout this chapter.

We also ran the behavioural anchor suite from Section 3.2.1 on both non-production backbones. Qwen3.6 scores 0.816 {[}0.650, 0.982{]} and gpt-oss-20b scores 0.647 {[}0.452, 0.842{]}, with the latter figure already reflecting the two harness modifications described in Section 6.1. The confidence intervals overlap, so the gap between the two models is not as big as it appears. However, 0.647 falls below the suite\textquotesingle s written passing threshold, which means that a group adopting a backbone from a new model family should expect the router configuration itself to require adjustment, not only the underlying model.

\subsection*{5.5 Retrieval on external anchors}\label{retrieval-on-external-anchors}

Query variants were generated once with the Qwen3.6-era prompt template and frozen before scoring, so all retrievers are evaluated on identical inputs regardless of which model would have issued them in production. We use two external benchmarks: SciFact from BEIR, whose 5,183-document collection and single-gold-passage claim-verification format make it a clean validation target for the measurement pipeline, and LitSearch, a 64,183-paper corpus that approximates the scale and query diversity of the retrieval workload the system faces in production.

\textbf{Table 5.9 SciFact (BEIR), nDCG@10.}

{\small
\begin{longtable}[]{@{}
  >{\raggedright\arraybackslash}p{(\columnwidth - 2\tabcolsep) * \real{0.5000}}
  >{\raggedright\arraybackslash}p{(\columnwidth - 2\tabcolsep) * \real{0.5000}}@{}}
\toprule\noalign{}
\begin{minipage}[b]{\linewidth}\raggedright
\textbf{Retriever}
\end{minipage} & \begin{minipage}[b]{\linewidth}\raggedright
\textbf{nDCG@10 {[}95\% CI{]}}
\end{minipage} \\
\begin{minipage}[b]{\linewidth}\raggedright
BM25
\end{minipage} & \begin{minipage}[b]{\linewidth}\raggedright
0.652 {[}0.61, 0.70{]}
\end{minipage} \\
\begin{minipage}[b]{\linewidth}\raggedright
SPECTER-dense
\end{minipage} & \begin{minipage}[b]{\linewidth}\raggedright
0.479 {[}0.43, 0.53{]}
\end{minipage} \\
\begin{minipage}[b]{\linewidth}\raggedright
RRF {[}BM25, SPECTER{]}
\end{minipage} & \begin{minipage}[b]{\linewidth}\raggedright
0.621 {[}0.58, 0.67{]}
\end{minipage} \\
\begin{minipage}[b]{\linewidth}\raggedright
SPECTER-dense, canonical {[}SEP{]} format
\end{minipage} & \begin{minipage}[b]{\linewidth}\raggedright
0.494 (interval not recorded)
\end{minipage} \\
\midrule\noalign{}
\endhead
\bottomrule\noalign{}
\endlastfoot
\end{longtable}
}

On the 300 SciFact queries, BM25 achieves an nDCG@10 of 0.652 (95\% CI {[}0.61, 0.70{]}), within 0.013 of the published BEIR value of 0.665 {[}5{]}. SPECTER-dense, a dense encoder pre-trained on scientific publication data, scores 0.479 ({[}0.43, 0.53{]}), well below BM25 (p \textless{} 0.001). Reciprocal rank fusion of the two scores 0.621, statistically indistinguishable from BM25 alone. When one component is substantially weaker than the other, fusing it dilutes the ranking rather than complementing it, and the fused list lands between the two individual scores.

\textbf{Table 5.10 LitSearch, 597 queries, 64,183-paper corpus.}

{\small
\begin{longtable}[]{@{}
  >{\raggedright\arraybackslash}p{(\columnwidth - 8\tabcolsep) * \real{0.2000}}
  >{\raggedright\arraybackslash}p{(\columnwidth - 8\tabcolsep) * \real{0.2000}}
  >{\raggedright\arraybackslash}p{(\columnwidth - 8\tabcolsep) * \real{0.2000}}
  >{\raggedright\arraybackslash}p{(\columnwidth - 8\tabcolsep) * \real{0.2000}}
  >{\raggedright\arraybackslash}p{(\columnwidth - 8\tabcolsep) * \real{0.2000}}@{}}
\toprule\noalign{}
\begin{minipage}[b]{\linewidth}\raggedright
\textbf{Retriever}
\end{minipage} & \begin{minipage}[b]{\linewidth}\raggedright
\textbf{nDCG@10 {[}95\% CI{]}}
\end{minipage} & \begin{minipage}[b]{\linewidth}\raggedright
\textbf{R@10}
\end{minipage} & \begin{minipage}[b]{\linewidth}\raggedright
\textbf{R@100}
\end{minipage} & \begin{minipage}[b]{\linewidth}\raggedright
\textbf{MRR}
\end{minipage} \\
\begin{minipage}[b]{\linewidth}\raggedright
BM25
\end{minipage} & \begin{minipage}[b]{\linewidth}\raggedright
0.378 {[}0.345, 0.413{]}
\end{minipage} & \begin{minipage}[b]{\linewidth}\raggedright
0.511
\end{minipage} & \begin{minipage}[b]{\linewidth}\raggedright
0.699
\end{minipage} & \begin{minipage}[b]{\linewidth}\raggedright
0.349
\end{minipage} \\
\begin{minipage}[b]{\linewidth}\raggedright
\textbf{BGE-dense (production)}
\end{minipage} & \begin{minipage}[b]{\linewidth}\raggedright
\textbf{0.485 {[}0.453, 0.516{]}}
\end{minipage} & \begin{minipage}[b]{\linewidth}\raggedright
0.637
\end{minipage} & \begin{minipage}[b]{\linewidth}\raggedright
0.829
\end{minipage} & \begin{minipage}[b]{\linewidth}\raggedright
0.451
\end{minipage} \\
\begin{minipage}[b]{\linewidth}\raggedright
Citation re-rank 0.7/0.3
\end{minipage} & \begin{minipage}[b]{\linewidth}\raggedright
0.469 {[}0.437, 0.503{]}
\end{minipage} & \begin{minipage}[b]{\linewidth}\raggedright
0.609
\end{minipage} & \begin{minipage}[b]{\linewidth}\raggedright
0.829
\end{minipage} & \begin{minipage}[b]{\linewidth}\raggedright
0.442
\end{minipage} \\
\begin{minipage}[b]{\linewidth}\raggedright
RRF {[}BM25, BGE{]}
\end{minipage} & \begin{minipage}[b]{\linewidth}\raggedright
0.490 {[}0.455, 0.526{]}
\end{minipage} & \begin{minipage}[b]{\linewidth}\raggedright
0.628
\end{minipage} & \begin{minipage}[b]{\linewidth}\raggedright
0.830
\end{minipage} & \begin{minipage}[b]{\linewidth}\raggedright
0.462
\end{minipage} \\
\midrule\noalign{}
\endhead
\bottomrule\noalign{}
\endlastfoot
\end{longtable}
}

On the 597-query LitSearch set the ordering inverts. BGE-dense, the production encoder, scores 0.485 nDCG@10 against 0.378 for BM25 (+0.107, p \textless{} 0.001), and the gap persists at depth: R@10 rises from 0.511 to 0.637 and R@100 from 0.699 to 0.829. A citation-graph re-ranker then adjusts the BGE top-100 candidate list by weighting each document\textquotesingle s final score as 0.7 times its semantic similarity plus 0.3 times its normalised in-corpus citation count. This shifts rank positions toward heavily cited papers at the expense of semantically similar but less-cited ones, costing 0.016 nDCG@10 relative to unreranked BGE (p = 0.014). In this corpus, being cited often does not make a paper more likely to answer a specific query; the re-ranker promotes documents that are prominent in the field rather than those that are relevant to the question asked. RRF between BM25 and BGE scores 0.490, statistically indistinguishable from BGE alone, confirming the SciFact observation that fusing a second retriever into an already-strong ranking adds no measurable gain.

One domain caveat applies to these rows: the queries cover machine learning and natural language processing topics, whereas the production corpus serves chemistry, biophysics, and membrane biology, so the relative ordering of retrievers observed here may not replicate in the other domains. The encoder bake-off in Table 5.1 addresses this directly by embedding a fixed pool of 190 in-domain sources and 5,000 distractors once under each candidate encoder, allowing all models to be compared on identical inputs drawn from the production corpus without a full re-embedding pass per model.

\subsection*{5.6 Reliability and cost}\label{reliability-and-cost}

Table 5.11 reports tool-use data for the agentic arm on each backbone, alongside the wall-clock cost of all three arms. A tool failure is a call that returned an error or a payload the model could not use. The recovery rate is the fraction of queries that encountered at least one such failure and still produced a final answer.

\textbf{Table 5.11 Tool-use reliability and per-query cost, agentic arm.} Error and degraded rates are omitted because the definition of a web-fetch failure changed between the Qwen runs; calls per query and recovery rate are comparable across all three.

{\small
\begin{longtable}[]{@{}
  >{\raggedright\arraybackslash}p{(\columnwidth - 6\tabcolsep) * \real{0.2500}}
  >{\raggedright\arraybackslash}p{(\columnwidth - 6\tabcolsep) * \real{0.2500}}
  >{\raggedright\arraybackslash}p{(\columnwidth - 6\tabcolsep) * \real{0.2500}}
  >{\raggedright\arraybackslash}p{(\columnwidth - 6\tabcolsep) * \real{0.2500}}@{}}
\toprule\noalign{}
\begin{minipage}[b]{\linewidth}\raggedright
\end{minipage} & \begin{minipage}[b]{\linewidth}\raggedright
\textbf{Qwen3.6}
\end{minipage} & \begin{minipage}[b]{\linewidth}\raggedright
\textbf{Qwen3.8}
\end{minipage} & \begin{minipage}[b]{\linewidth}\raggedright
\textbf{gpt-oss-20b}
\end{minipage} \\
\begin{minipage}[b]{\linewidth}\raggedright
Tool calls, total
\end{minipage} & \begin{minipage}[b]{\linewidth}\raggedright
1,458
\end{minipage} & \begin{minipage}[b]{\linewidth}\raggedright
1,380
\end{minipage} & \begin{minipage}[b]{\linewidth}\raggedright
1,742
\end{minipage} \\
\begin{minipage}[b]{\linewidth}\raggedright
Calls per query
\end{minipage} & \begin{minipage}[b]{\linewidth}\raggedright
7.33
\end{minipage} & \begin{minipage}[b]{\linewidth}\raggedright
6.93
\end{minipage} & \begin{minipage}[b]{\linewidth}\raggedright
8.75
\end{minipage} \\
\begin{minipage}[b]{\linewidth}\raggedright
Queries with at least one tool failure
\end{minipage} & \begin{minipage}[b]{\linewidth}\raggedright
84
\end{minipage} & \begin{minipage}[b]{\linewidth}\raggedright
102
\end{minipage} & \begin{minipage}[b]{\linewidth}\raggedright
58
\end{minipage} \\
\begin{minipage}[b]{\linewidth}\raggedright
\textbf{Recovery rate}
\end{minipage} & \begin{minipage}[b]{\linewidth}\raggedright
\textbf{1.000}
\end{minipage} & \begin{minipage}[b]{\linewidth}\raggedright
\textbf{1.000}
\end{minipage} & \begin{minipage}[b]{\linewidth}\raggedright
\textbf{1.000}
\end{minipage} \\
\begin{minipage}[b]{\linewidth}\raggedright
Wall-clock per query, agentic, egress = full
\end{minipage} & \begin{minipage}[b]{\linewidth}\raggedright
67.0 s
\end{minipage} & \begin{minipage}[b]{\linewidth}\raggedright
157.3 s
\end{minipage} & \begin{minipage}[b]{\linewidth}\raggedright
29.3 s
\end{minipage} \\
\begin{minipage}[b]{\linewidth}\raggedright
Wall-clock per query, agentic, egress = off
\end{minipage} & \begin{minipage}[b]{\linewidth}\raggedright
54.1 s
\end{minipage} & \begin{minipage}[b]{\linewidth}\raggedright
96.0 s
\end{minipage} & \begin{minipage}[b]{\linewidth}\raggedright
21.3 s
\end{minipage} \\
\begin{minipage}[b]{\linewidth}\raggedright
Wall-clock per query, bare
\end{minipage} & \begin{minipage}[b]{\linewidth}\raggedright
15.7 s
\end{minipage} & \begin{minipage}[b]{\linewidth}\raggedright
8.0 s
\end{minipage} & \begin{minipage}[b]{\linewidth}\raggedright
2.4 s
\end{minipage} \\
\begin{minipage}[b]{\linewidth}\raggedright
Agentic : bare
\end{minipage} & \begin{minipage}[b]{\linewidth}\raggedright
4.3×
\end{minipage} & \begin{minipage}[b]{\linewidth}\raggedright
\textasciitilde20×
\end{minipage} & \begin{minipage}[b]{\linewidth}\raggedright
\textasciitilde12×
\end{minipage} \\
\begin{minipage}[b]{\linewidth}\raggedright
Serving profile
\end{minipage} & \begin{minipage}[b]{\linewidth}\raggedright
one GPU, alone
\end{minipage} & \begin{minipage}[b]{\linewidth}\raggedright
two GPUs, tensor-parallel, shared with users
\end{minipage} & \begin{minipage}[b]{\linewidth}\raggedright
one GPU, alone, MXFP4
\end{minipage} \\
\begin{minipage}[b]{\linewidth}\raggedright
Recovery rate, egress = off
\end{minipage} & \begin{minipage}[b]{\linewidth}\raggedright
1.000
\end{minipage} & \begin{minipage}[b]{\linewidth}\raggedright
1.000
\end{minipage} & \begin{minipage}[b]{\linewidth}\raggedright
0.965
\end{minipage} \\
\midrule\noalign{}
\endhead
\bottomrule\noalign{}
\endlastfoot
\end{longtable}
}

Failed tool calls do not cause the harness to abandon a query. Across all three backbones, and including the corpus-only arms in which external retrieval is disabled, every question that encountered at least one tool call returning an error or an unusable payload still produced a final answer by the end of the run. When a call fails, the harness either retries the same call, substitutes a different retrieval tier, or proceeds to compose an answer from the evidence it has already gathered. In none of the observed runs did it abort the conversation or return a partial result. (recovery rate of 1.000 in Table 5.11 for every arm of every backbone at egress = full). The corpus-only arms also function as a natural test of the egress guard. With outbound requests disabled at the network layer, the models continued to issue web-search and Semantic Scholar calls during the course of a run, indicating that the instruction to use those tools was still followed. Each such call was intercepted and refused by the per-request guard before it could leave the machine, so no external traffic was generated while the full sequence of attempted tool calls was preserved in the log. Because the guard rejects each blocked call with a standard error response rather than crashing the process, the model can respond to it in the same way as any other tool failure. The recovery behaviour is therefore almost identical to the unrestricted arm: 1.000 on both Qwen backbones, and 0.965 on gpt-oss-20b, where two of the 57 queries that encountered a refused call ended without a final answer. This is the only figure below 1.000 in the entire suite.

The three backbones differ more in which tools they call than in how reliably those tools respond. gpt-oss-20b keeps 85\% of its tool calls inside the local corpus search ladder and escalates to web search on only 17 of the 199 questions. The two Qwen backbones utilise the external tier more, issuing 331 (Qwen3.8) and 362 (Qwen3.6) web searches over the same set respectively. This is consistent with the egress decomposition in Section 5.1, where the external retrieval tiers contribute +0.080 to accuracy on gpt-oss but +0.166 (Qwen3.8) and +0.211 (Qwen3.6) on the two Qwen models. A model that rarely leaves the corpus has correspondingly less to gain from the tiers beyond it.

However, a high count of web searches shows only that the model looks outward. It does not show that the final answer actually depends on the material retrieved. The corpus-only arm answers that question directly by disabling the external tiers entirely. On Qwen3.8, more than half of the total harness improvement remains when all outbound requests are blocked, indicating that a substantial share of the supporting evidence for each answer comes from the local corpus even in the configuration where the model also searches the web.

When individual retrieval tiers are degraded by injected faults, the harness does not stop or return a degraded answer. Instead it increases the number of tool calls per query from a baseline of 8.6 to between 13.9 and 16.1, routing around the failed tier through alternate sources. The grounding score remains within 0.04 of its fault-free value, and every query still produces a final answer.

Wall-clock cost depends primarily on the backbone and its serving configuration rather than on the harness itself, so the cost rows across columns should not be compared directly. The Qwen3.8 arms ran on the production service alongside other users, while the Qwen3.6 and gpt-oss arms ran on dedicated hardware with no competing load. The agentic loop takes approximately four times as long as a bare model call on Qwen3.6, twelve times as long on gpt-oss-20b, and twenty times as long on the dense Qwen3.8 configuration, at a mean of seven to nine tool calls per query.

\subsection*{5.7 Comparison with published baselines}\label{comparison-with-published-baselines}

Table 5.12 compares the harness against the published LitQA2 baselines. The Munin rows are drawn from the standalone answer track, whose protocol matches the published benchmark. A single question is asked, the model produces a single answer, with a deadline of 900 seconds. For completeness, the agentic arm from the ablation in Section 5.1 is also listed alongside. The two protocols yield consistent results on every backbone. On Qwen3.8 the question-paired difference between the standalone and the agentic arm is +0.010 {[}-0.035, +0.055{]}, and the two arms agree on the outcome for 175 of the 199 questions. On Qwen3.6 the two accuracies are 0.874 and 0.869; on gpt-oss they are 0.528 and 0.563. Across the two Qwen backbones themselves, the standalone-track accuracies differ by +0.010 {[}-0.030, +0.050{]} (p = 0.78), again with 181 of 199 verdicts in agreement.

\textbf{Table 5.12 LitQA2 accuracy and precision for each arm.} Accuracy counts abstentions and unparseable responses as incorrect; precision is computed only over questions on which the model produced an answer, matching the definitions used in the published benchmarks.

{\small
\begin{longtable}[]{@{}
  >{\raggedright\arraybackslash}p{(\columnwidth - 6\tabcolsep) * \real{0.2500}}
  >{\raggedright\arraybackslash}p{(\columnwidth - 6\tabcolsep) * \real{0.2500}}
  >{\raggedright\arraybackslash}p{(\columnwidth - 6\tabcolsep) * \real{0.2500}}
  >{\raggedright\arraybackslash}p{(\columnwidth - 6\tabcolsep) * \real{0.2500}}@{}}
\toprule\noalign{}
\begin{minipage}[b]{\linewidth}\raggedright
\textbf{System}
\end{minipage} & \begin{minipage}[b]{\linewidth}\raggedright
\textbf{Accuracy}
\end{minipage} & \begin{minipage}[b]{\linewidth}\raggedright
\textbf{Precision}
\end{minipage} & \begin{minipage}[b]{\linewidth}\raggedright
\textbf{Note}
\end{minipage} \\
\begin{minipage}[b]{\linewidth}\raggedright
PaperQA2 (Skarlinski et al.~2024)
\end{minipage} & \begin{minipage}[b]{\linewidth}\raggedright
0.660 ± 0.012
\end{minipage} & \begin{minipage}[b]{\linewidth}\raggedright
0.852 ± 0.011
\end{minipage} & \begin{minipage}[b]{\linewidth}\raggedright
trained on LitQA2 splits
\end{minipage} \\
\begin{minipage}[b]{\linewidth}\raggedright
Human experts (n = 9)
\end{minipage} & \begin{minipage}[b]{\linewidth}\raggedright
0.677 ± 0.119
\end{minipage} & \begin{minipage}[b]{\linewidth}\raggedright
0.738 ± 0.096
\end{minipage} & \begin{minipage}[b]{\linewidth}\raggedright
\end{minipage} \\
\begin{minipage}[b]{\linewidth}\raggedright
Munin, Qwen3.6-35B-A3B
\end{minipage} & \begin{minipage}[b]{\linewidth}\raggedright
0.874 {[}0.824, 0.920{]}
\end{minipage} & \begin{minipage}[b]{\linewidth}\raggedright
0.951 {[}0.918, 0.978{]}
\end{minipage} & \begin{minipage}[b]{\linewidth}\raggedright
ablation arm 0.869 / 0.930
\end{minipage} \\
\begin{minipage}[b]{\linewidth}\raggedright
\textbf{Munin, Qwen3.8-27B}
\end{minipage} & \begin{minipage}[b]{\linewidth}\raggedright
\textbf{0.884 {[}0.839, 0.925{]}}
\end{minipage} & \begin{minipage}[b]{\linewidth}\raggedright
\textbf{0.926 {[}0.889, 0.963{]}}
\end{minipage} & \begin{minipage}[b]{\linewidth}\raggedright
ablation arm 0.874 / 0.946
\end{minipage} \\
\begin{minipage}[b]{\linewidth}\raggedright
Munin, gpt-oss-20b
\end{minipage} & \begin{minipage}[b]{\linewidth}\raggedright
0.528 {[}0.462, 0.593{]}
\end{minipage} & \begin{minipage}[b]{\linewidth}\raggedright
0.847 {[}0.774, 0.911{]}
\end{minipage} & \begin{minipage}[b]{\linewidth}\raggedright
ablation arm 0.563 / 0.896; 11 empty answers
\end{minipage} \\
\midrule\noalign{}
\endhead
\bottomrule\noalign{}
\endlastfoot
\end{longtable}
}

We need to keep three things in mind when we compare Munin with the published baselines. First, PaperQA2 was trained on the LitQA2 training and evaluation splits. The models reported here have no such exposure, so a slightly worse performance would have been expected. Second, the Munin rows cover only the 199 questions whose source papers are present in the local corpus. A question whose source is absent measures the system\textquotesingle s abstention behaviour rather than its answering ability, which is the subject of Section 5.3 and is handled separately. Third, run-to-run variability on a 199-question arm is approximately ±0.035, so the two Qwen rows should be treated as a single value rather than two independent estimates.

Both Qwen backbones exceed the published PaperQA2 accuracy by approximately 0.22 and the human-expert baseline by approximately 0.20, while also producing a higher precision than either. The figure most directly relevant to the premise of this paper is however not in the table: when outbound requests are disabled entirely so that no request leaves the institution, the corpus-only harness scores 0.704 on Qwen3.6 and 0.663 on Qwen3.8 at a precision above 0.91. That result still exceeds the published PaperQA2 accuracy, which means the gap over the prior best system is attributable to the retrieval-and-verification loop operating over a local corpus rather than to access to external indices. gpt-oss-20b, the single-GPU backbone, reaches an accuracy of 0.528 on this benchmark, with 11 of its 199 responses being empty turn-ends that carry no content. On this set it trails PaperQA2 on accuracy but matches it on precision.

\section*{6 Discussion}\label{discussion}

\subsection*{6.1 Model-agnosticism and re-certification}\label{model-agnosticism-and-re-certification}

A stronger base model does not guarantee improved performance when the surrounding pipeline was tuned to a specific model. AstaBench {[}22{]} reports that substituting a frontier model into existing agent configurations produced modest gains on a generic agent while degrading several specialised agents below their prior scores, indicating that the interaction between model and harness is not monotone in model capability.

For a retrieval framework, model-agnosticism must therefore be established empirically for each model rather than inferred from the existence of a compatible API. Which stages of the pipeline are affected by a model swap, and by how much, depend on how the new model handles the tool-calling protocol, formats structured outputs, and calibrates its refusal threshold. The appropriate response to a model upgrade is to re-run the evaluation suite and record where the operating points move.

During development, each milestone was checked against the frozen evaluation suite before deployment. This process surfaced three defects that would not have been visible from accuracy alone: two affected the system itself and one was an error in the evaluation harness. The full account of these issues is given in Appendix B. We ship the suite with the framework so that an adopting group can re-certify the pipeline on its own base model rather than relying on our reported scores. The frozen question sets make this a single overnight run per model, comparable in wall-clock time to one benchmark sweep.

Adding a different model from the same model family required no changes to the harness, but adding a different model did. The first pass of gpt-oss-20b through the full ablation scored an accuracy of 0.467, well below its final value of 0.563. Two parsing defects were responsible. First, the serving stack attached channel-identifier tokens to roughly three per cent of tool names in the prompt, which caused the model to issue calls with malformed arguments. Second, the frozen scorer recognised the "Answer: A" response format used by the Qwen models but not the bolded "Answer: A" format that gpt-oss produces, so correct answers were counted as unparseable. Both fixes were applied at the parser and scorer boundaries and did not alter any Qwen verdict. The pre-repair scorecard is retained in the repository alongside the corrected one in the supplementary material so that the cost of these mismatches is visible to anyone reproducing the result.

This means that a backbone swap across model families is first a test of the tool contract between the harness and the model, and only secondarily a test of the model\textquotesingle s ability to answer questions. Any component of the harness that interprets the model\textquotesingle s output format, including the call parser, the answer scorer, and the router\textquotesingle s example set, should be expected to require inspection before the first number produced by the new backbone can be trusted.

\subsection*{6.2 Limitations}\label{limitations}

We want to clarify that this paper introduces no new model, embedding, or retrieval algorithm. Its contribution is the integration of existing components into a deployable system and the evaluation suite that characterises how each component contributes. The results are based on three models from two developers. This shows that the harness effect is not an artefact of a single model, but it does not establish that the effect will hold for any future model, and the magnitude of the gain is already model-dependent: on gpt-oss-20b it is roughly one third of the Qwen figure.

The harness is also not entirely model-agnostic. For each family we integrated, a small number of adjustments were required. Three modifications addressed clarification-handling behaviour specific to the Qwen models. Two adjusted routing gates for the Qwen3.6 configuration, which does not expose a reasoning-effort parameter, and one repaired a tool-name mismatch in the call parser, paired with a matching update to the answer scorer, to handle the response format used by gpt-oss. When we state that the harness generalises, it still requires minor, documented, per-family adjustments so that the code runs.

We want to state two other limitations. First, "agentic" refers specifically to this harness, the particular set of tools and control flow described in Section 3. The ablation in Section 5.1 compares this one harness against two baselines; it does not compare the class of agentic systems against the class of retrieval-augmented systems. Second, the certification thresholds that ship with the evaluation suite are regression gates over committed baselines. They flag when a successive run has drifted outside a pre-set range relative to its predecessor, but they do not predict whether any individual answer is correct or trustworthy. We also state one other practical limitation. We have not yet tested this on models that fit on a 16 GB GPU, instead opting for gpt-oss-20b which at the production context window requires 24 GB of VRAM, which we still find acceptable for a research group.

Several design choices limit how far the numbers in Section 5 can be generalised. All answering results use the 199 LitQA2 questions whose source papers were ingested into the corpus, not the full benchmark. These questions are multiple-choice and a substantial fraction are answerable from parametric model knowledge alone, which is why the bare arm already reaches 0.387 and why the corpus-only control is necessary to separate retrieval-driven gains from those the model could produce without any document access. However, PaperQA2 was trained on the LitQA2 training and evaluation splits while the models reported here had no such exposure, and the human-expert figure it sits beside is a mean over nine experts with a standard deviation of 0.119.

gpt-oss-20b differs from the Qwen backbones simultaneously in developer, architecture, quantisation, and sampling configuration, so the smaller harness effect observed on this model cannot be attributed to any single cause from the data collected here. In addition, the bare and RAG arms sample at temperature 0.7 while the agentic arm uses the research profile settings, an asymmetry that applies to every run reported. Repeated 199-question arms differ by up to 0.035 in accuracy, so paired differences below approximately three points throughout Section 5 fall within the measurement floor. The correct-abstention rates in Section 5.3.2 are computed on subsets of 20 to 32 answerable questions, which gives wide confidence intervals. The observed cross-model agreement is consistent with both true equality and modest differences, and should not be read as strong evidence for either. Finally, cost is reported as wall-clock time and token count rather than GPU-seconds, because per-job accounting was not deployed on the production cluster during the runs reported here.

The faithfulness judge evaluates each claim against a single passage, so claims supported by multiple passages jointly receive no credit. Every faithfulness figure is therefore a lower bound. The judge was validated on RAGTruth, a general-domain benchmark, and applied here to scientific answers without in-domain validation. A second-judge cross-check would have required sending evaluation data off-site, which the deployment rules do not permit.

Abstention is detected by a text marker in the answer. On Qwen3.8, 13 of the 100 correct refusals of fabricated papers also cited a real corpus document alongside the abstention marker. These are scored as correct abstentions because the marker is present, though only four were read manually to determine whether the real paper was offered as related work or presented as the answer. As the suite stands, retrieving the wrong paper from the corpus is invisible to the scorer.

gpt-oss-20b ends its turn without a final message on several arms (25 bare, 162 RAG, 20 C1). The protocol scores these as wrong for all backbones alike, so its bare and RAG accuracies and C1 refusal rate are floors. Precision over attempted answers is the fairer summary for those arms, and its faithfulness rests on the 21 questions both arms answered. The risk-coverage figures are single operating points because the harness emits a binary abstain decision with no confidence score, so no curve can be drawn.

None of the public benchmarks used in this paper are overly similar to the domain we deploy Munin in. SciFact covers biomedical claim verification, LitSearch is drawn from machine-learning literature, and LitQA2 is a biology benchmark, whereas the production corpus is chemistry, biophysics, and membrane biology. The retrieval anchors in Section 5.5 additionally use query variants that were generated by a single expander model.

The sovereignty claim also has some caveats. Inference, the corpus, all indexes, the sandbox, and the evaluation judge run on the group\textquotesingle s own hardware. However, chat traffic transits the rented web host as stated in Section 3.5, and at the full egress level the web-search tool sends queries to a commercial search API. This is not inherently violating sovereignty, but in our deployment some data still leaves the system. The privacy of other deployments relies on the operators of those systems. All numbers in this paper come from one deployment and one corpus in our research consortium. The corpus itself is not published, so the reported figures can be reproduced in method only. Every statistic can be recomputed from the per-question verdicts in the released scorecards. Because every answering result is computed on a public benchmark\textquotesingle s question set, evaluation against the group\textquotesingle s own queries is deferred until a curated local query pool has been assembled.

\section*{7 Conclusion}\label{conclusion}

Groups that want a language model grounded in their own literature have so far had to choose between a hosted service that can access the documents but receives unpublished work as input, deploy a self-hosted chat frontend that keeps the data local but gives the model no way to retrieve, read, or check its answers against those documents, or build the full retrieval-and-verification stack in-house from components that have no standard interface between them. Munin is an open-source framework that attempts to make self-hosted deployment much easier. It indexes a group\textquotesingle s own corpus with its citation graph and wraps whichever open-weight model the group can serve on its existing hardware in an agent harness that provides retrieval, full-text reading, and citation-verification tools. The contribution of the paper is the claim that this harness, rather than the specific model inside it, is what makes the system useful on domain literature, and the measurement apparatus that tests that claim ships alongside the software so that any installing group can re-run the evaluation on its own corpus and model.

On the reference deployment, the harness approximately doubles the base model\textquotesingle s accuracy on literature questions and also doubles the fraction of answer claims that the model can trace to a specific passage it read. It holds attempted answers to the same evidence standard across a backbone swap, and declines to answer rather than fabricate when the corpus does not contain the source document, without producing a single invented local citation across three backbones and three hundred fabricated references.

The harness improves both models from the Qwen family, as well as gpt-oss-20b, the latter running on a single consumer-grade GPU. The gain comes from the retrieval-and-verification loop rather than from any particular model. With all outbound requests blocked and the system consulting only the local corpus, both Qwen models still exceed the strongest published result on the benchmark (0.704 and 0.663 against 0.660), while gpt-oss-20b does not.

The evaluation suite ships with the framework, so an adopting group runs it on its own corpus and model rather than relying on the numbers reported here. Re-running the suite after each model swap or harness change gives the group a running record of how its system performs over time, which is the information it needs to decide when a new backbone is worth deploying.

\subsection*{Availability of code, data and scorecards}\label{availability-of-code-data-and-scorecards}

Munin is released under the Apache-2.0 licence at https://github.com/elgetilab/munin. The version evaluated in this paper is tagged v1.0.0 and archived at https://doi.org/10.5281/zenodo.23191404. The release contains the evaluation suite, which re-runs as one command, together with the comparison and certification commands, the 54 scorecards with per-question verdicts and faithfulness scores for every arm on every backbone, the frozen question sets (the 199 LitQA2 in-corpus questions, the C1 fabricated-paper set, the C2b shadow-corpus set with its removal list), the routing anchor suite and the RAGTruth validation set. Model weights are the public releases listed in Table 4.2 with their revision hashes. The reference corpus consists of the research group\textquotesingle s licensed articles and is not redistributed. The LitQA2 source papers were retrieved from their public DOIs and added to that corpus. A group following the same procedure on its own corpus will obtain comparable rather than identical figures, because retrieval difficulty, the citation graph and the shadow-corpus outcomes depend on the remaining corpus.

\subsection*{Acknowledgements}\label{acknowledgements}

We thank Alexander Vogel for his advice on cluster configuration and Jan-Ingo Haas for fruitful discussions about LLMs. Funded by the Deutsche Forschungsgemeinschaft (DFG, German Research Foundation), project number 514664767 (subprojects B07 and INF, to M.E.).

\subsection*{Author contributions}\label{author-contributions}

Author contributions (CRediT). Markus Fischer: conceptualisation, methodology, software, investigation, formal analysis, data curation, validation, visualisation, resources, writing (original draft), writing (review and editing). Jens Meiler: funding acquisition, resources. Matthias Elgeti: conceptualisation, supervision, funding acquisition, resources, project administration, writing (original draft), writing (review and editing).

\subsection*{Competing interests}\label{competing-interests}

The authors declare no competing interests.

\subsection*{Declaration of AI use}\label{declaration-of-ai-use}

Large language models were used in this work in four declared roles. First, as the object of study: the three backbones of Table 4.2 run inside the harness and in the baseline arms, with per-question verdicts released in the scorecards. Second, as an evaluation instrument: MiniCheck-Flan-T5-Large, running on premises, scores claim-level faithfulness in Section 5.2 after validation on RAGTruth (Section 4.5); no hosted model was used as a judge. Third, in developing the software: the Munin codebase and evaluation suite were written with the assistance of Anthropic\textquotesingle s Claude Opus (4.8, 5.0 and 5.5) and of Munin itself (Qwen3.6-35B-A3B and Qwen3.8-27B); coding agents also executed evaluation runs and produced provenance reports from the repository, such as the commit audit behind the egress caveat in Section 6.2. Every result in Section 5 is computed by the committed evaluation code from the recorded model outputs, and every reported number was checked by the authors against the scorecards. Fourth, in preparing the manuscript: Munin running Qwen3.8-27B was used for most rewriting and editing, and Claude Opus (5.0 and 5.5) for drafting and restructuring parts of Sections 4 and 6 and for checking numerical consistency between the text and the scorecards. All text was reviewed and revised by the authors, who take full responsibility for the content.

\section*{References}\label{references}
\begin{list}{}{\setlength{\leftmargin}{2.2em}\setlength{\labelwidth}{2em}\setlength{\itemsep}{2pt}\small\sloppy}
\item[{[}1{]}] M. D. Skarlinski, S. Cox, J. M. Laurent, J. D. Braza, M. Hinks, M. J. Hammerling, M. Ponnapati, S. G. Rodriques, A. D. White. Language agents achieve superhuman synthesis of scientific knowledge. arXiv:2409.13740v2, 2024.
\item[{[}2{]}] J. Lála, O. O\textquotesingle Donoghue, A. Shtedritski, S. Cox, S. G. Rodriques, A. D. White. PaperQA: Retrieval-augmented generative agent for scientific research. arXiv:2312.07559, 2023.
\item[{[}3{]}] S. Narayanan, J. D. Braza, R.-R. Griffiths, M. Ponnapati, A. Bou, J. Laurent, O. Kabeli, G. Wellawatte, S. Cox, S. G. Rodriques, A. D. White. Aviary: Training language agents on challenging scientific tasks. arXiv:2412.21154, 2024.
\item[{[}4{]}] A. Ajith, M. Xia, A. Chevalier, T. Goyal, D. Chen, T. Gao. LitSearch: A retrieval benchmark for scientific literature search. EMNLP 2024. arXiv:2407.18940.
\item[{[}5{]}] N. Thakur, N. Reimers, A. Rücklé, A. Srivastava, I. Gurevych. BEIR: A heterogeneous benchmark for zero-shot evaluation of information retrieval models. NeurIPS 2021 Datasets and Benchmarks. arXiv:2104.08663.
\item[{[}6{]}] D. Wadden, S. Lin, K. Lo, L. L. Wang, M. van Zuylen, A. Cohan, H. Hajishirzi. Fact or fiction: Verifying scientific claims. EMNLP 2020.
\item[{[}7{]}] L. Tang, P. Laban, G. Durrett. MiniCheck: Efficient fact-checking of LLMs on grounding documents. EMNLP 2024. arXiv:2404.10774.
\item[{[}8{]}] C. Niu, Y. Wu, J. Zhu, S. Xu, K. Shum, R. Zhong, J. Song, T. Zhang. RAGTruth: A hallucination corpus for developing trustworthy retrieval-augmented language models. ACL 2024, pp. 10862--10878. arXiv:2401.00396.
\item[{[}9{]}] J. Foo, P. S. Prasad, S. Khoo. Know or not: A library for evaluating out-of-knowledge base robustness. arXiv:2505.13545, 2025.
\item[{[}10{]}] H. Joren, J. Zhang, C.-S. Ferng, D.-C. Juan, A. Taly, C. Rashtchian. Sufficient context: A new lens on retrieval augmented generation systems. ICLR 2025. arXiv:2411.06037.
\item[{[}11{]}] A. Muhamed, L. F. R. Ribeiro, M. Dreyer, V. Smith, M. T. Diab. RefusalBench: Generative evaluation of selective refusal in grounded language models. arXiv:2510.10390, 2025.
\item[{[}12{]}] P. Kirichenko, M. Ibrahim, K. Chaudhuri, S. J. Bell. AbstentionBench: Reasoning LLMs fail on unanswerable questions. NeurIPS 2025 Datasets and Benchmarks. arXiv:2506.09038.
\item[{[}13{]}] B. Wen, J. Yao, S. Feng, C. Xu, Y. Tsvetkov, B. Howe, L. L. Wang. Know your limits: A survey of abstention in large language models. TACL 13:529--556, 2025. arXiv:2407.18418.
\item[{[}14{]}] S. Machcha, S. Yerra, S. Gupta, A. Sahoo, S. Sultana, H. Yu, Z. Yao. Knowing when to abstain: Medical LLMs under clinical uncertainty. EACL 2026, pp. 6153--6182. arXiv:2601.12471.
\item[{[}15{]}] J. Chen, H. Lin, X. Han, L. Sun. Benchmarking large language models in retrieval-augmented generation. AAAI 2024, 38(16):17754--17762. arXiv:2309.01431.
\item[{[}16{]}] A. T. Kalai, O. Nachum, S. S. Vempala, E. Zhang. Why language models hallucinate. arXiv:2509.04664, 2025.
\item[{[}17{]}] Z. Yuan et al. CiteAudit: You cited it, but did you read it? A benchmark for verifying scientific references in the LLM era. arXiv:2602.23452, 2026.
\item[{[}18{]}] CiteCheck: Retrieval-grounded detection of LLM citation hallucinations in scientific text. arXiv:2605.27700, 2026.
\item[{[}19{]}] M. Li, Z. Lin, S. Ma. Source or it didn\textquotesingle t happen: A multi-agent framework for citation hallucination detection. arXiv:2605.08583, 2026.
\item[{[}20{]}] D. Rao, C. Callison-Burch. BibTeX citation hallucinations in scientific publishing agents: Evaluation and mitigation. arXiv:2604.03159, 2026.
\item[{[}21{]}] LLM hallucinations in the wild: Large-scale evidence from non-existent citations. arXiv:2605.07723, 2026.
\item[{[}22{]}] Allen Institute for AI. AstaBench: Rigorous benchmarking of AI agents with a scientific research suite. arXiv:2510.21652, 2025.
\item[{[}23{]}] D. Wang, M. Cheng, Q. Liu, S. Yu, Z. Liu, Z. Guo. PaperArena: An evaluation benchmark for tool-augmented agentic reasoning on scientific literature. arXiv:2510.10909, 2025.
\item[{[}24{]}] L. Xiong, K. Luo, Z. Xia, W. Zhang, J.-G. Yao, Z. Liu et al. AutoResearchBench: Benchmarking AI agents on complex scientific literature discovery. arXiv:2604.25256, 2026.
\item[{[}25{]}] Y. Bang, Z. Ji, A. Schelten, A. Hartshorn, T. Fowler, C. Zhang, N. Cancedda, P. Fung. HalluLens: LLM hallucination benchmark. arXiv:2504.17550, 2025.
\item[{[}26{]}] F. S. Bao, M. Li, R. Qu, G. Luo et al. FaithBench: A diverse hallucination benchmark for summarization by modern LLMs. NAACL 2025, pp. 448--461.
\item[{[}27{]}] S. Mysore, T. O\textquotesingle Gorman, A. McCallum, H. Zamani. CSFCube: A test collection of computer science research articles for faceted query by example. NeurIPS 2021 Datasets and Benchmarks. arXiv:2103.12906.
\item[{[}28{]}] A. Jacovi et al. The FACTS Grounding leaderboard: Benchmarking LLMs\textquotesingle{} ability to ground responses to long-form input. arXiv:2501.03200, 2025.
\item[{[}29{]}] F. Wang et al. Astute RAG: Overcoming imperfect retrieval augmentation and knowledge conflicts for large language models. arXiv:2410.07176, 2024.
\item[{[}30{]}] Resisting contextual interference in RAG via parametric-knowledge reinforcement. arXiv:2506.05154, 2025.
\item[{[}31{]}] When retrieval doesn\textquotesingle t help: A large-scale study of biomedical RAG. arXiv:2606.04127, 2026.
\item[{[}32{]}] SoK: The attack surface of agentic AI --- tools and autonomy. arXiv:2603.22928, 2026.
\item[{[}33{]}] Acceptance-test-driven evaluation protocols for business-centric LLM systems. arXiv:2606.02755, 2026.
\item[{[}34{]}] Who judges the judges? Governance from metrics: A runtime framework for continuous LLM compliance monitoring. arXiv:2605.24737, 2026.
\item[{[}35{]}] Toward pre-deployment assurance for enterprise AI agents: Ontology-grounded simulation and trust certification. arXiv:2606.04037, 2026.
\item[{[}36{]}] Beyond static leaderboards: Predictive validity for the evaluation of LLM agents. arXiv:2606.19704, 2026.
\item[{[}37{]}] PredictaBoard: Benchmarking LLM score predictability. arXiv:2502.14445, 2025.
\item[{[}38{]}] A. Singh, J. C. Chang, C. Anastasiades, D. Haddad, A. Naik, A. Tanaka, A. Zamarron, C. Nguyen, J. D. Hwang, J. Dunkleberger, M. Latzke, S. Rao, J. Lochner, R. Evans, R. Kinney, D. S. Weld, D. Downey, S. Feldman. Ai2 Scholar QA: Organized literature synthesis with attribution. arXiv:2504.10861, 2025.
\item[{[}39{]}] A. Asai, J. He, R. Shao, W. Shi, A. Singh, J. C. Chang, K. Lo, L. Soldaini, S. Feldman, M. D\textquotesingle Arcy, D. Wadden, M. Latzke, M. Tian, P. Ji, S. Liu, H. Tong, B. Wu, Y. Xiong, L. Zettlemoyer, G. Neubig, D. S. Weld, D. Downey, W. Yih, P. W. Koh, H. Hajishirzi. OpenScholar: Synthesizing scientific literature with retrieval-augmented LMs. arXiv:2411.14199, 2024.
\item[{[}40{]}] Open WebUI contributors. Open WebUI. Software, https://github.com/open-webui/open-webui, accessed 2026-09-24.
\item[{[}41{]}] D. Avila and LibreChat contributors. LibreChat. Software, https://github.com/danny-avila/LibreChat, accessed 2026-09-24.
\item[{[}42{]}] Mintplex Labs. AnythingLLM. Software, https://github.com/Mintplex-Labs/anything-llm, accessed 2026-09-24.
\end{list}

\clearpage
\section*{Appendix A Tool Inventory}\label{appendix-a-tool-inventory}

The system exposes four agents and 41 standalone tools (Table A1). An agent wraps its tool calls in a multi-step reasoning loop; a standalone tool is invoked directly by the outer model, executes one deterministic step, and returns. Most operations in the system fall into the second category, and wrapping them in an agent layer would add no capability, only latency.

\textbf{Table A1. Full tool inventory.}

{\small
\begin{longtable}[]{@{}
  >{\raggedright\arraybackslash}p{(\columnwidth - 2\tabcolsep) * \real{0.2000}}
  >{\raggedright\arraybackslash}p{(\columnwidth - 2\tabcolsep) * \real{0.8000}}@{}}
\toprule\noalign{}
\begin{minipage}[b]{\linewidth}\raggedright
\textbf{Category}
\end{minipage} & \begin{minipage}[b]{\linewidth}\raggedright
\textbf{Tools}
\end{minipage} \\
\begin{minipage}[b]{\linewidth}\raggedright
Agents
\end{minipage} & \begin{minipage}[b]{\linewidth}\raggedright
source, search, compute, deep\_research
\end{minipage} \\
\begin{minipage}[b]{\linewidth}\raggedright
Retrieval
\end{minipage} & \begin{minipage}[b]{\linewidth}\raggedright
paper\_search, semantic\_scholar\_search, paper\_lookup, web\_search, web\_fetch, search\_user\_docs, list\_documents, browse\_tag\_papers, check\_papers\_availability, get\_paper\_pdf
\end{minipage} \\
\begin{minipage}[b]{\linewidth}\raggedright
Citation graph
\end{minipage} & \begin{minipage}[b]{\linewidth}\raggedright
get\_citations, get\_references, s2\_get\_citations, s2\_get\_references, get\_author\_papers, export\_citations
\end{minipage} \\
\begin{minipage}[b]{\linewidth}\raggedright
LLM utility
\end{minipage} & \begin{minipage}[b]{\linewidth}\raggedright
llm\_summarize, transcribe\_equation, view\_attachment
\end{minipage} \\
\begin{minipage}[b]{\linewidth}\raggedright
Artifacts
\end{minipage} & \begin{minipage}[b]{\linewidth}\raggedright
create\_artifact, read\_artifact, update\_artifact, list\_artifacts, save\_artifact\_to\_documents
\end{minipage} \\
\begin{minipage}[b]{\linewidth}\raggedright
Memory
\end{minipage} & \begin{minipage}[b]{\linewidth}\raggedright
remember, forget, recall, search\_past\_conversations
\end{minipage} \\
\begin{minipage}[b]{\linewidth}\raggedright
Projects
\end{minipage} & \begin{minipage}[b]{\linewidth}\raggedright
list\_projects, get\_current\_project
\end{minipage} \\
\begin{minipage}[b]{\linewidth}\raggedright
Execution
\end{minipage} & \begin{minipage}[b]{\linewidth}\raggedright
run\_python, edit\_python, sandbox\_reset, compile\_latex, calculate
\end{minipage} \\
\begin{minipage}[b]{\linewidth}\raggedright
Orchestration
\end{minipage} & \begin{minipage}[b]{\linewidth}\raggedright
invoke\_agent, tool\_search, ask\_clarification, faq
\end{minipage} \\
\begin{minipage}[b]{\linewidth}\raggedright
Planning
\end{minipage} & \begin{minipage}[b]{\linewidth}\raggedright
set\_plan, update\_plan\_item
\end{minipage} \\
\midrule\noalign{}
\endhead
\bottomrule\noalign{}
\endlastfoot
\end{longtable}
}

Tools run in parallel by default; eleven carry an explicit serialisation flag and execute one at a time. These are the state mutators: five artifact writers, three memory writers, the sandbox reset, and the plan setter. The remaining 34 tools are read-only or idempotent, so the flag is needed on a small, known set. The default-permit choice is deliberate. If a new tool is added without the flag and it is a true mutator, the race produces a visible duplicate write or lost update in the next integration test. Under a default-forbid policy the same omission would silently serialise the tool and slow the system with no error to investigate.

\section*{Appendix B: Design rationale}\label{appendix-b-design-rationale}

This appendix documents the measurement results during development that led to each architectural choice in the current system. Entries follow the order in which the decisions were made.

\subsubsection*{B.1 Encoder migration}\label{b.1-encoder-migration}

Replacing SPECTER-dense with BGE-large raised Recall@10 on the local pool from 0.44 to 0.73 (+0.29, p \textless{} 0.001), but end-to-end answer accuracy over the same period increased by only 0.075. Because the correct paper was present in the candidate set even before this change, the additional recall did not translate into more correct answers. We therefore stopped further optimisation of the retrieval layer and directed subsequent effort at the reading step.

\subsubsection*{B.2 Removal of passage summarisation}\label{b.2-removal-of-passage-summarisation}

The original implementation produced a summary of each retrieved passage before passing it to the answering model, to reduce token expenditure. The summary removed the numerical values that the questions required. In the evaluation this appeared as an increase in abstentions rather than incorrect answers, because the model correctly identified that the needed information was not present in its context. When we supplied unsummarised full text in an oracle run, accuracy rose to 0.82 and the pinned full-text reading agent (source) achieved 0.864 under production conditions. We removed the summarisation step. The reading agent now receives the raw tagged reference text, and the associated token cost is accepted as necessary for preserving the evidence the model requires.

\subsubsection*{B.3 Introduction of the agentic loop}\label{b.3-introduction-of-the-agentic-loop}

Prepending the top-5 retrieved documents to every query produced lower accuracy than the bare model on both backbones (Section 5.1.1). The fixed context directed the model toward passages that did not contain the answer, and in doing so reduced its willingness to attempt an answer from parametric knowledge. We replaced the fixed-context approach with the iterative agentic loop, in which the model selects what to retrieve, determines how many documents to pull, decides whether to use the result, and chooses when to stop searching.

\subsubsection*{B.4 Guardrail implementation via code counters}\label{b.4-guardrail-implementation-via-code-counters}

Two prompt-level interventions intended to constrain tool-use behaviour produced no measurable effect or a negative one. A research fragment encouraging more thorough investigation raised the median tool-call count per query from 10.5 to 20.5 without changing accuracy. A second prompt modification had reversed itself within two runs. The intervention that worked was a hard counter in the execution loop: setting CHAT\_MAX\_TOOL\_CALLS to 30 reduced the maximum call count from 43 to 31 and the 90th percentile from 40 to 30, while grounding scores remained within noise. We therefore implement all behavioural guardrails as code-level constraints rather than prompt instructions.

\subsubsection*{B.5 Pre-deployment bug detection by the evaluation suite}\label{b.5-pre-deployment-bug-detection-by-the-evaluation-suite}

A controlled A/B comparison on LitSearch revealed a score-scale mismatch in the citation re-ranker: documents scored under different normalisation bases were compared directly, compressing the ranking distribution. nDCG@10 was 0.117 before the fix and 0.469 after. The C2b paired shadow-corpus test showed that correct abstention on missing-source questions improved from 0.200 to 0.667 across the harness upgrade that introduced explicit not-in-corpus detection (+0.467, p \textless{} 0.001). Neither defect would have been apparent from a single-query demonstration or from accuracy measurements alone. We ship the evaluation suite with the framework so that an adopting group can verify these fixes on its own corpus and detect regressions introduced by subsequent upgrades.

\subsubsection*{B.6 A defect of the faithfulness suite}\label{b.6-a-defect-of-the-faithfulness-suite}

The first capture run for the faithfulness evaluation enumerated retrieval tools from a list that predated two later stages of the pipeline: the corpus search ladder and the grounded-read step. As a result, the judge scored the agentic arm\textquotesingle s claims against only a fraction of the passages the model had actually read. The paired faithfulness difference came out as essentially zero (+0.010 to +0.023), and because the bug was in the capture step rather than in the model, the null replicated across two backbones and three runs. Re-running the capture with the complete tool list gave +0.258 on Qwen3.8 and +0.336 on Qwen3.6 (both p \textless{} 0.001), which are the figures reported in Section 5.2. The lesson is that a result replicating across backbones can replicate an upstream error just as readily as a real effect.

\section*{Appendix C Metric definitions}\label{appendix-c-metric-definitions}

\textbf{Retrieval.} A ranked list is the document identifiers in rank order, best first, with repeated identifiers counted only at their first occurrence following TREC convention. Relevance grades are taken from the benchmark\textquotesingle s qrels as integers, and documents not listed receive grade 0. nDCG@k uses linear gain, meaning the raw grade value, discounted by log2(rank + 1) at 1-indexed ranks, and the ideal DCG is computed from the qrels grades themselves rather than from the ranked list. Recall@k is the fraction of relevant documents appearing in the top k results, and MRR is the reciprocal rank of the first relevant document, averaged over all queries.

\textbf{Answering.} Each response is classified into one of four categories: correct, incorrect, abstain, or unparseable. Accuracy is defined as correct responses over all questions, so both abstentions and unparseable outputs count against the arm. The term "answered" refers to responses in which the model committed to a specific option, whether correct or incorrect. Precision of attempted is the fraction of answered responses that are correct, coverage is the fraction of all questions that were answered, and selective risk is the fraction of answered responses that are incorrect, which equals one minus precision of attempted. A risk-coverage operating point is simply the pair (coverage, risk) for one arm on one population. The unparseable count is reported separately for each arm so that an arm producing many malformed responses cannot silently exclude them from its denominator.

\textbf{Faithfulness.} An answer is decomposed into individual claims, and the judge scores each claim against the union of all passages the arm retrieved on that turn. The per-answer score is the fraction of claims judged supported, and the headline figure is the mean of these per-answer scores across the arm\textquotesingle s answers, so that long and short answers contribute equally regardless of how many claims they contain. Paired differences between arms are computed only on the questions that both arms answered.

\textbf{Intervals and tests.} Confidence intervals are 95\% percentile bootstraps resampling query indices with a fixed seed, using 1,000 resamples for the ablation comparisons and 2,000 for the abstention tests. Paired comparisons report both a two-sided paired bootstrap p-value and a Wilcoxon signed-rank test applied to the same question set in the same order. For proportions computed on small answerable subsets of 20 to 32 items, where a percentile bootstrap can only land on multiples of 1/n and degenerates near zero or one, we also compute the closed-form Wilson score interval and quote whichever is wider when the two disagree. A third interval, which resamples the full paired question set and re-derives the answerable subset within each resample, is stored in the scorecards for transparency but not quoted in the main text.

\end{document}